\documentclass[runningheads]{llncs}

\usepackage{accv}

\usepackage{accvabbrv}

\usepackage{graphicx}
\usepackage{booktabs}

\usepackage[accsupp]{axessibility}  

\usepackage[pagebackref,breaklinks,colorlinks,citecolor=accvblue]{hyperref}

\usepackage{orcidlink}

\usepackage{comment}
\usepackage{multirow}
\usepackage{multicol}
\usepackage{float}
\DeclareMathOperator*{\argmax}{argmax}
\usepackage{bbm}  
\usepackage[normalem]{ulem}
\usepackage{tabularx}
\usepackage[export]{adjustbox}

\newcommand{\methodname}{TranSPORTmer}
\usepackage[dvipsnames]{xcolor}

\begin{document}

\title{TranSPORTmer: A Holistic Approach to Trajectory Understanding in Multi-Agent Sports} 

\titlerunning{TranSPORTmer}

\author{Guillem Capellera\inst{1,2}\orcidlink{0009-0006-7266-078X} \and
Luis Ferraz\inst{1}\orcidlink{0000-0001-7851-9193} \and Antonio Rubio\inst{1}\orcidlink{0000-0002-6771-8645} \and Antonio Agudo\inst{2}\orcidlink{0000-0001-6845-4998} \and Francesc Moreno-Noguer\inst{2}\orcidlink{0000-0002-8640-684X}}

\authorrunning{G.~Capellera {\em et al.}}

\institute{Kognia Sports Intelligence, Barcelona, Spain \\ \email{\{guillem.capellera,luis.ferraz,antonio.rubio\}@kogniasports.com} \and 
Institut de Robòtica i Informàtica Industrial CSIC-UPC, Barcelona, Spain
\email{\{gcapellera,aagudo,fmoreno\}@iri.upc.edu} }

\maketitle

\begin{abstract}
  Understanding trajectories in multi-agent scenarios requires addressing various tasks, including predicting future movements, imputing missing observations, inferring the status of unseen agents, and classifying different global states. Traditional data-driven approaches often handle these tasks separately with specialized models. We introduce TranSPORTmer, a unified transformer-based framework capable of addressing all these tasks, showcasing its application to the intricate dynamics of multi-agent sports scenarios like soccer and basketball. Using Set Attention Blocks, TranSPORTmer effectively captures temporal dynamics and social interactions in an equivariant manner. The model's tasks are guided by an input mask that conceals missing or yet-to-be-predicted observations. Additionally, we introduce a CLS extra agent to classify states along soccer trajectories, including \textit{passes}, \textit{possessions}, \textit{uncontrolled} states, and \textit{out-of-play} intervals, contributing to an enhancement in modeling trajectories. Evaluations on soccer and basketball datasets show that TranSPORTmer outperforms state-of-the-art task-specific models in player forecasting, player forecasting-imputation, ball inference, and ball imputation. \href{https://youtu.be/8VtSRm8oGoE}{https://youtu.be/8VtSRm8oGoE}
  \keywords{Multi-agent modelling \and Imputation \and Transformers.}
\end{abstract}

\section{Introduction}
\label{sec:intro}

\begin{figure}[t]
    \centering
    \includegraphics[width=1.0\textwidth]{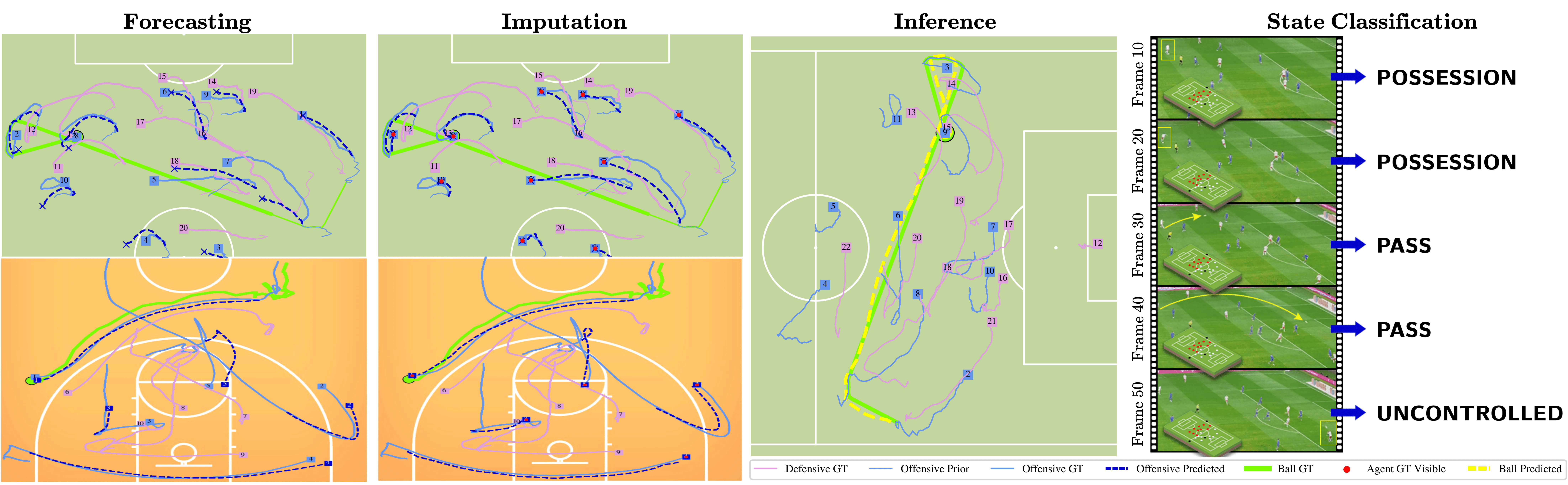}
    \captionof{figure}{\textbf{\methodname~} is a holistic model that is able to perform multiple tasks for trajectory understanding in multi-agent sport scenarios. The images showcase examples using soccer and basketball data for the tasks of  {\em forecasting}: predicting future trajectories given past observations; {\em imputation}: predicting agent trajectories given partial observations; {\em inference}: predicting the trajectory of an unobserved agent given the state of other ones;  and {\em state classification}: assigning a semantic label to each frame of the sequence.  Continuous and dashed lines correspond to observed states and predicted trajectories, respectively.}
    \label{fig:FigRecap_horizontal}
\end{figure}

Multi-agent systems are prevalent in various real-world scenarios encompassing pedestrian modelling \cite{alahi2016social,gupta2018social, amirian2019social, kosaraju2019social,salzmann2020trajectron++,ngiam2021scene,girgis2021latent, navarro2022social, saadatnejad2023social, xu2023eqmotion}, human pose estimation \cite{fragkiadaki2015recurrent,jain2016structural,martinez2017human,mao2019learning,mao2020history,aksan2021spatio,cai2020learning,guo2023back}, and sports analytics~\cite{zheng2016generating, zhan2018generating, hauri2021multi, alcorn2021baller2vec++, hu2022entry}. This paper focuses on the latter, where trajectory understanding plays a pivotal role in unraveling the inter-dependencies within multi-agent sports scenarios. This understanding opens up diverse applications such as  performance evaluation \cite{decroos2019actions, brito2020association, teranishi2022evaluation}, scouting \cite{pappalardo2019playerank}, tactical analysis \cite{decroos2018automatic, wang2023tacticai} and event detection \cite{fassmeyer2021toward,vidal2022automatic}. In contrast to urban contexts, the realm of sports requires the capturing of both individual player influences and comprehensive team strategies, all of which involve heightened levels of interactions and complex dynamics.

Despite the promising applications, challenges persist in this domain. Inherent inaccuracies in optical tracking data, often arising from occlusions, pose a significant hurdle. The substantial costs associated with adopting GPS technology for ball tracking~\cite{kim2023ball} add an extra layer of complexity. Additionally, the intricacies introduced by off-screen players~\cite{omidshafiei2022multiagent, xu2023uncovering} and the nuances of broadcasting videos further contribute to the challenges in multi-agent sports scenarios. Moreover, the annotation of a match demands a significant amount of manual work due to the density of events and states that unfold during gameplay.

Previous research has proposed task-specific solutions for trajectory forecasting \cite{yeh2019diverse, ding2020graph, capellera2024footbots} and imputing missing observations \cite{omidshafiei2022multiagent, liu2019naomi}. Some works offer unified frameworks capable of addressing both tasks \cite{qi2020imitative, xu2023uncovering}. However, a common limitation across these models is the assumption that all agents have either complete or partially observed data, overlooking scenarios involving entirely unseen agents. Furthermore, several of these models rely on recursive prediction strategies, potentially compromising efficiency in match processing and performance when modeling long-range sequences.

In the domain of unseen agent inference, recent efforts have concentrated on predicting both ball location \cite{kim2023ball} and player positions \cite{everett2023inferring}. Nevertheless, these approaches require additional data beyond agent locations, including velocities and customized event data. Moreover, prior works focusing on event and state classification using trajectory data often center around a limited set of sparse scenarios \cite{fassmeyer2021toward} or specific events like passes and receptions \cite{kim2023ball, honda2022pass}, without providing comprehensive annotation for every state of the game.

In this paper, we present \methodname, a comprehensive approach for trajectory understanding in multi-agent sports scenarios. Our approach uses transformer encoders, or Set Attention Blocks (SABs)~\cite{lee2019set, devlin2018bert, dosovitskiy2020image, kong2020sound}, to capture temporal dynamics and inter-agent (or ``social'') interactions, maintaining agent permutation equivariance. To enhance adaptability, we use a socio-temporal mask for handling missing or future observations and defining game tasks like predicting opponent movements. Building on CLS tokens~\cite{devlin2018bert}, we introduce the CLS extra agent for state classification at each timestep alongside trajectory completion tasks. We also implement a learnable uncertainty mask in the loss function to improve predictions near visible observations by modeling their uncertainty. Our method is validated on one soccer and two basketball datasets.
\noindent The key contributions can be summarized as follows:
\begin{itemize}
    \item We develop a holistic transformer-based model that integrates trajectory forecasting, imputation, inference, and state classification in multi-agent sports scenarios, outperforming state-of-the-art task-specific methods.
    \item We propose a CLS extra agent to infer per-frame game states, achieving robust state classification while enhancing trajectory completion accuracy.
    \item We implement a learnable uncertainty mask in the loss function for boundary observations, which reflects uncertainty and leads to more accurate predictions.
    \item We analyze the coarse-to-fine manner of our architecture in the ball inference task, resulting in a 25\% improvement over current state-of-the-art methods.
\end{itemize}
\section{Related Work}
\label{sec:related}
This section discusses the related work in trajectory forecasting, imputation, inference, and state classification, with a specific emphasis on multi-agent sports scenarios.

\noindent\textbf{Trajectory Forecasting} consists in predicting future positions based on past observations. In the context of multi-agent sports, earlier approaches~\cite{felsen2018will, zhan2018generating, zheng2016generating}  predicted long-term behaviors using Variational Recurrent Neural Networks (VRNNs)~\cite{chung2015recurrent}. However, these methods lack equivariance properties and rely on heuristics like tree-based role alignment~\cite{lucey2013representing, sha2017fine} to define a specific {\em ordering} of the agents. The combination of VRNNs with Graph Neural Networks (GNNs)~\cite{battaglia2018relational}, results in GVRNN~\cite{yeh2019diverse, sun2019stochastic}, defining an equivariant model treating agents as nodes of a fully connected graph. This approach allows the aggregation of spatial interactions for final predictions. However, GVRNN treat agent dependencies equally by aggregating agent information at each timestep. To handle dependencies between different agents more effectively, \cite{huang2019stgat,ding2020graph,monti2021dag,bertugli2021ac, fassmeyer2022semi} used a Graph Attention Network (GAT)~\cite{velivckovic2017graph}, replacing fully connected graphs. Transformer-based models~\cite{vaswani2017attention} have been used in this task~\cite{alcorn2021baller2vec, alcorn2021baller2vec++}, demonstrating a superior performance compared to graph-recurrent-based methods. Nevertheless, conducting attention in both temporal and social dimensions simultaneously still incurs a notable computational cost. In contrast, \methodname~employs attention in both temporal and social dimensions sequentially. This design choice results in a substantial reduction in computational cost without compromising performance. Moreover, by departing from recursive sequence construction, our model gains a significant advantage in long-term sequence prediction, thanks to its inherent look-ahead temporal property.

\noindent\textbf{Trajectory Imputation}  involves predicting agents' behavior in unknown frames using available information, such as partial trajectories of the target agent. Previous research tackled value imputation in time series with an autoregressive RNN~\cite{cao2018brits}. A bidirectional GVRNN structure proposed by~\cite{omidshafiei2022multiagent} addressed imputing missing agent observations in soccer games. However, due to its autoregressive nature, these approaches may lead to suboptimal results in long-range sequences~\cite{gu2017non, lee2018deterministic}. Liu {\em et al.}~\cite{liu2019naomi} introduced a non-autoregressive imputation model exploiting the multi-resolution structure of sequential data, although it falls short in handling trajectory forecasting. Another asynchronous approach solved imputation and forecasting tasks using imitative techniques~\cite{qi2020imitative}. Some research leveraged GVRNN to handle both tasks simultaneously~\cite{xu2023uncovering}. Similarly, our method is equipped to handle this unified task effectively.


\noindent \textbf{Trajectory Inference}  aims to predict the behavior of agents across all frames based solely on information from other agents. This   is often approached as ball inference~\cite{amirli2022prediction, kim2023ball}. The fusion of Set Transformers~\cite{lee2019set}  with Bi-LSTM~\cite{hochreiter1997long} has been utilized to infer the ball trajectory and identify the ball possessor (or pass receiver). This method relies on player trajectories and their corresponding velocities~\cite{kim2023ball}. As we will demonstrate later, \methodname~does not require player velocities to infer the ball position. Moreover, it can be applied  to any type of agent, including the goalkeeper, that exhibits very particular motion patterns.


\noindent\textbf{State/Event Classification:} On this context,  ~\cite{vidal2022automatic} applied a rule-based framework to identify soccer events based on agent trajectories. \cite{fassmeyer2021toward} proposed a method using a variational autoencoder and support vector machine to detect events such as corner kicks, crosses, and counterattacks. Another significant work is~\cite{honda2022pass}, introducing a pass receiver Transformer-LSTM model that integrates visual information with player and ball trajectories. The recent work previously mentioned for ball inference~\cite{kim2023ball} can also serve as a pass receiver prediction model. However, these approaches primarily focus on limited soccer context situations such as set pieces and often rely on robust and precise estimations of ball and/or player trajectories. \methodname~provides a more detailed coverage of events, referred to as states, including \textit{passes}, \textit{possessions}, \textit{uncontrolled} situations, and transitions between in-play and \textit{out-of-play} states. The model also demonstrates robustness against missing observations, showcasing its ability to perform state classification even with an unseen ball.

\noindent\textbf{Pedestrian Motion Modeling:}
We review advances in pedestrian motion modeling, noting that Becker \etal~\cite{becker2018red} found an RNN with an MLP decoder outperformed social pooling methods~\cite{alahi2016social, lee2017desire, gupta2018social} despite lacking social encoding. Transformer-based models have also advanced the field, with \cite{giuliari2021transformer} achieving strong results on the TrajNet benchmark~\cite{sadeghian2018trajnet} by focusing on temporal dynamics. Subsequent approaches~\cite{girgis2021latent, aksan2021spatio} improved social interaction modeling using transformer encoders without positional encoding. Recently, diffusion models~\cite{rempe2023trace, xie2023omnicontrol, mao2023leapfrog} have emerged for stochastic human behavior modeling.

\section{Revisiting Attention Mechanisms}
\label{sec:background}
Attention mechanisms are effective at capturing relationships in sequences or sets. Utilizing $n$ queries $\mathbf{Q}$ and $n_v$ keys $\mathbf{K}$ of dimension $d_k$, and $n_v$ values $\mathbf{V}$ of dimension $d_v$ as inputs, the attention mechanism computes weighted sums of values by assessing the compatibility between queries and keys  measured using dot or scaled dot products. The masked attention expression $\text{A}(\mathbf{Q}, \mathbf{K}, \mathbf{V}, \mathbf{M})$, incorporating a binary mask $\mathbf{M}$, can be written as: 
\begin{equation}
\text{softmax}\left(\frac{(\mathbf{Q}\mathbf{K}^{\top}) + o(\mathbf{M})}{\sqrt{d_k}}\right)\mathbf{V},
\end{equation}
with $\mathbf{Q} \in \mathbb{R}^{n \times d_k}$, $\mathbf{K} \in \mathbb{R}^{n_v \times d_k}$, $\mathbf{V} \in \mathbb{R}^{n_v \times d_v}$, and $\mathbf{M} \in \{0, 1\}^{n \times n_v}$. $\mathbf{M}$ determines which keys are used in computing attention for each query. Specifically, entries filled with zeros in $\mathbf{M}$ indicate keys to be included, while entries filled with ones denote those to be excluded. The function $o(\cdot)$ maps 0/1 values to 0/$-\infty$. Note that the softmax operator output will assign zero weight to the latter set of keys, ensuring that similarity scores are normalized. The weighted value sum is obtained by multiplying attention weights with their corresponding values.

In practice, attention mechanism is often extended with multiple attention heads, also called \textit{Multi-Head Attention} (MHA)~\cite{vaswani2017attention}, allowing to capture different aspects of the data. Instead of computing a single attention function, this method projects $\mathbf{Q}$, $\mathbf{K}$, $\mathbf{V}$ onto $H$ different $d^{h}_k$, $d^{h}_k$, $d^{h}_v$ dimensional vectors, respectively. Each attention head computes its own attention weights and weighted sum of values, and the outputs of the attention heads are concatenated or linearly transformed to obtain the final attention output.

$\text{MHA}(\mathbf{Q}, \mathbf{K}, \mathbf{V}, \mathbf{M})$ is inferred using $\text{concat}(\text{head}_1, \ldots, \text{head}_h, \ldots, \text{head}_H)\mathbf{W}^O$, where $\text{head}_h = \text{A}(\mathbf{Q}\mathbf{W}^Q_h, \mathbf{K}\mathbf{W}^K_h, \mathbf{V}\mathbf{W}^V_h, \mathbf{M})$ and $h=\{1,\ldots,H\}$ represents the $h$-th attention head. Therefore, $\text{MHA}$ is a $[n \times d]$ matrix, with  learnable parameters $\{\mathbf{W}^Q_h, \mathbf{W}^K_h\} \in \mathbb{R}^{d_k \times d^h_k}$, $\mathbf{W}^V_h \in \mathbb{R}^{d_v \times d^h_v}$, and $\mathbf{W}^O \in \mathbb{R}^{hd^h_v \times d}$.  In this work we will use $d_k = d_v = d$ and $d^h_k=d^h_v=d_k/H$ as it is standard in literature.

The MHA operation was extended~\cite{lee2019set} to operate on sets by defining the SAB. Given one set of $d$-dimensional vectors and one mask determining which vectors are used to compute the attention, denoted by $\mathbf{X}$ and $\mathbf{M}$, respectively, the SAB is defined as:
\begin{equation}
    \text{SAB}(\mathbf{X}, \mathbf{M}) = \text{LayerNorm}(\mathbf{H} + \text{rFFN}(\mathbf{H})), 
\end{equation}
where $\mathbf{H} = \text{LayerNorm}(\mathbf{X} + \text{MHA}(\mathbf{X}, \mathbf{X}, \mathbf{X}, \mathbf{M}))$ and $\text{rFFN}(\mathbf{H})$ denotes the row-wise feed-forward neural network applied to $\mathbf{H}$.
Note that SAB is an adaptation of the encoder block of the transformer but lacks the positional encoding. The MHA operation itself provides the property of permutation equivariance, allowing SAB to effectively capture relationships in the absence of positional information. When no mask is provided, the SAB operation is denoted as $\text{SAB}(\mathbf{X})=\text{SAB}(\mathbf{X}, \mathbf{0})$, where $\mathbf{0}$ denotes an all-zero-values mask.


\section{Method}
\label{sec:method}

\subsection{Problem Statement}

Let us consider a  set of $N \in \mathbb{N}$ agent observations by including players and a ball in our case, denoted as $X = \{\mathbf{x}^1, \ldots, \mathbf{x}^n, \ldots, \mathbf{x}^N\}$ with $n=\{1,\ldots,N\}$, where each observation contains the $(x,y)$ pitch locations. We can now collect $T$ observations along time for every agent, defining the tensor $\mathbf{X}_{1:T}$ where all $\mathbf{x}^{n}_{t}$ with $t=\{1,\ldots,T\}$ are considered. Trajectory completion aims at inferring missing or unobserved entries of a data structure based on the visible ones. Given partial observations $\mathbf{X}_{1:T}^{U}$ and a $[T \times N]$ binary mask $\mathbf{M}$ to encode by 0 the visible observations and by 1 the unobserved ones, the goal is to find a function $f_1(\cdot)$ to infer the full observations such that:
\begin{equation}
\label{completion}
   f_1(\mathbf{X}_{1:T}^{U}, \mathbf{M} ) = \mathbf{X}_{1:T}.
\end{equation}

Based on that idea, we can define three sub-tasks by imposing specific constraints on the mask $\mathbf{M}$:

\noindent\textbf{Trajectory forecasting:} Full observability is assumed for timesteps up to $\hat{t}<T$, with $\mathbf{M}$ entries for these observations set to 0.
    
\noindent\textbf{Trajectory imputation:} At least one observation per agent is available, meaning at least one null entry per row in $\mathbf{M}$.
    
\noindent\textbf{Trajectory inference:} This task is the most challenging, as it involves at least one agent having no observations throughout the entire duration, meaning that at least one entire row of the matrix $\mathbf{M}$ lacks null entries.

In addition, we delve into the classification of states within the game, seeking another function that takes the same input as in Eq.~\eqref{completion} but generates an output corresponding to a specific state for each timestep. These states involve the actions {\em pass}, {\em possession}, {\em uncontrolled}, and {\em out of play}, 4 in total, all of which are pertinent in a soccer context. Specifically, our objective is to estimate a classification function $f_2(\cdot)$ such that:
\begin{equation}
   f_2(\mathbf{X}_{1:T}^{U}, \mathbf{M} ) = \mathbf{s}_{1:T},
\end{equation}
where $\mathbf{s}_{1:T}$ is a [$T\times 4$] dimensional tensor that represents the probability distribution over each game state for each timestep.

\subsection{\methodname}

\begin{figure*}[t!]
  \centering
  \includegraphics[width=1.0\linewidth]{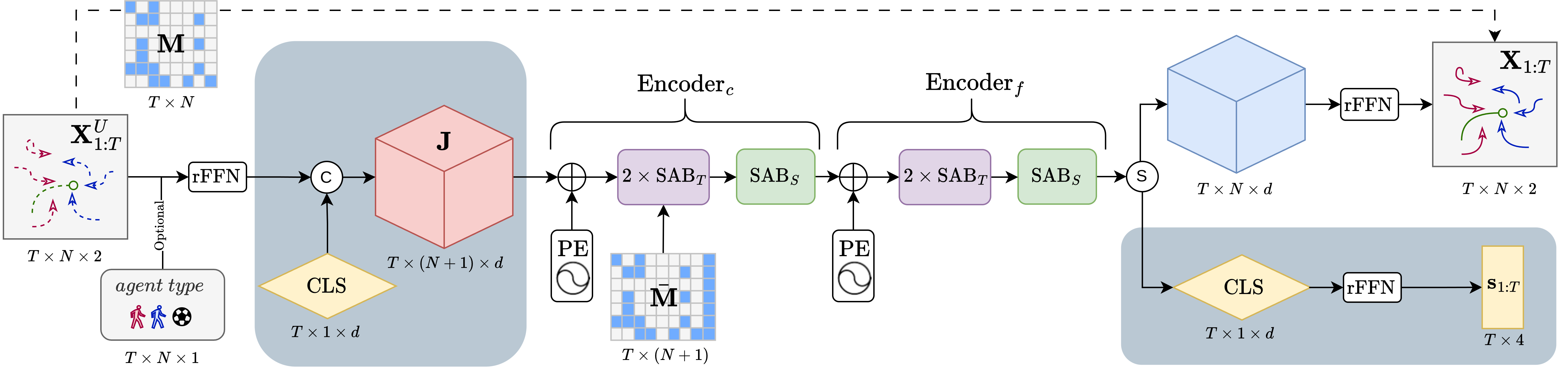}
\caption{\textbf{\methodname.} 
  The architecture uses sequential Set Attention Blocks for attention in both temporal (SAB$_T$) and social (SAB$_S$) axes. A Positional Encoder (PE) precedes each encoder to maintain the temporal sequence. The mask $\mathbf{M}$ identifies the values to be predicted (dashed arrow), forming the complete observation tensor $\mathbf{X}_{1:T}$.  The extended mask $\bar{\mathbf{M}}$ is applied to  the 2 $\times$ SAB$_T$ of the first Encoder$_c$, conveying information about hidden and visible states. Blue-gray segments are involved in state classification, including the CLS extra agent and the final classification head to rank the state classes per frame. (c) operation stands for concatenation and (s) for split.}
  \label{fig:TranSPORTmer}
\end{figure*}

We next present  TranSPORTmer our holistic and versatile approach to address trajectory forecasting, imputation, inference and state classification. \Cref{fig:TranSPORTmer} depicts its main components.

\noindent \textbf{Input processing:} The input tensor $\mathbf{X}_{1:T}^{U}$ contains partial observations, indicated by the input mask matrix $\mathbf{M}$. We can append additional known information to this tensor, such as the \textit{agent type}, which is an integer corresponding to each observation representing: 0 for ball, 1 for offensive team player, and 2 for defensive team player. Therefore, the shape of $\mathbf{X}_{1:T}^{U}$ can go from $[T \times M \times 2]$ to $[T \times M \times 3]$ with $(x,y,\textit{agent type})$. Initially, $\mathbf{X}_{1:T}^{U}$ is transformed by a row-wise feed-forward network (rFFN), becoming an embedding tensor of dimension $d$. 

\noindent \textbf{CLS extra agent:} Then a CLS tensor of dimension $[T\times d]$ is appended as an extra agent along the social axis,  resulting in a $[T \times (N+1) \times d]$ tensor $\mathbf{J}$. To ensure consistency, the mask $\bar{\mathbf{M}}$ of dimensions $[T \times (N+1)]$ extends $\mathbf{M}$, setting all entries corresponding to the CLS extra agent to one to indicate hidden observations. This extended mask is used in the initial SAB operations to make a first approximation of the hidden observations using temporal information.

\noindent \textbf{Coarse-to-fine encoders:} The  next block comprises two encoders, applied sequentially and that operate in a coarse-to-fine manner. Formally: 
\begin{align}
    \mathbf{J'} = \text{Encoder}_c (\mathbf{J},\bar{\mathbf{M}}) &= \text{SAB}_S \left( \text{SAB}_T \left( \text{SAB}_T (\mathbf{J} + \text{PE}, \bar{\mathbf{M}}), \bar{\mathbf{M}}\right) \right) , \\
    \text{Encoder}_f (\mathbf{J'}) &= \text{SAB}_S ( \text{SAB}_T ( \text{SAB}_T (\mathbf{J'} + \text{PE}) ) ) \;,
\end{align}
where $\text{PE}$ corresponds to the original positional encoder~\cite{vaswani2017attention} to preserve temporal ordering. SAB$_T$ and SAB$_S$ are temporal and social set attention blocks, respectively. SAB$_T$ processes individual temporal dynamics through the temporal embeddings of each agent, while SAB$_S$ addresses social interactions by encoding the embeddings of all agents at each timestep. The sequential configuration of SAB$_T$ followed by SAB$_S$ enables the implicit integration of information from both future and past time steps through temporal attention, enhancing the model’s ability to consider a broader temporal context in social attention.

\noindent \textbf{Output construction:} After passing through the encoder blocks, the output tensor retains the dimensions of $\mathbf{J}$. This output is then split into two tensors: the encoded trajectory embeddings and the encoded CLS extra agent. The former undergoes a rFFN operation to yield a tensor of dimension $[T \times N \times 2]$ corresponding to the predicted $(x,y)$ pitch locations. The binary mask $\mathbf{M}$ is then employed to directly propagate the visible $(x,y)$ values from the input tensor, $\mathbf{X}_{1:T}^{U}$, resulting in the full observation tensor $\mathbf{X}_{1:T}$. On the other side, the encoded CLS extra agent is reshaped and processed through another rFFN operation to obtain probability scores for each class at each time, $s_{1:T}$, of dimension $[T \times 4]$.

Note that our model exhibits permutation equivariance under agent permutations, as the operations along the social axis inherently maintain this property. In this architecture, an additional mask can be employed in all SAB blocks to ignore corrupt or NaN inputs observations not to predict, or to facilitate padding during batching with length-varying sequences and different numbers of agents involved. We denote this mask as NaN-mask.

\subsection{Loss Functions}
\begin{figure}[t]
    \centering
    \includegraphics[width=0.75\linewidth]{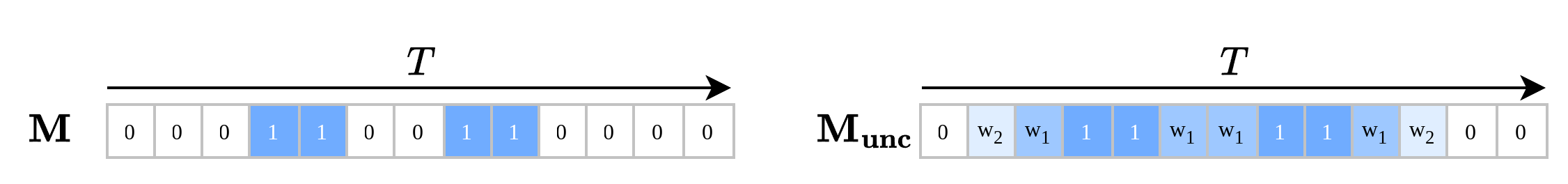}
    \caption{Binary mask ($\mathbf{M}$) and learnable uncertainty mask ($\mathbf{M_{unc}}$) for a single agent. Null values indicate visible observations.}
    \label{fig:munc}
\end{figure}

We introduce a learnable uncertainty mask $\mathbf{M_{unc}}$, with the same dimension as $\mathbf{M}$ to represent observation uncertainty. Here, $\mathbf{M_{unc}}^{n}_{t}=1$ where $\mathbf{M}^{n}_{t}=1$, indicating areas of maximum uncertainty (hidden observations). Along the time axis, we use two learnable weights: $w_1\in(0,1)$ bounded using a sigmoid function, and $w_2:=1-w_1$. These weights are applied to the immediate neighbors of 1's ($\mathbf{M_{unc}}^{n}_{t}=w_1$) and to the second neighbors if they are not immediate neighbors ($\mathbf{M_{unc}}^{n}_{t}=w_2$). All other values are set to null entries, signifying visible observations and, consequently, a lack of uncertainty. Extending the mask in the loss for boundary observations to reflect uncertainty enables the model to reconstruct them, leading to more accurate overall predictions. Figure~\ref{fig:munc} illustrates the differences between the binary mask ($\mathbf{M}$) and the learnable uncertainty mask ($\mathbf{M_{unc}}$) for a single agent over time.

The loss function for trajectory completion uses the Average Displacement Error (ADE) with the learnable uncertainty mask, assessing the disparity between the predictions and the ground truth:
\begin{equation}\label{eq:Loss}
    \mathcal{L}_\text{ADE} =(\sum_{n=1}^{N} \sum_{t=1}^{T}\mathbf{M_{unc}}_{t}^{n} )^{-1}\sum_{n=1}^{N} \sum_{t=1}^{T} \left\lVert \mathbf{\hat{x}}^{n}_{t} - \mathbf{x}^{n}_{t} \right\rVert_2 \mathbf{M_{unc}}_{t}^{n} \,,
\end{equation}
where $\mathbf{\hat{x}}^{n}_{t}$ denotes our estimation of the $n$-th agent at time $t$, $\mathbf{x}^{n}_{t}$ corresponds to the ground truth. We also utilize a standard Cross Entropy (CE) loss  as the training metric for the state classification task: 
\begin{equation}
\label{eq:CELoss}
 \mathcal{L}_\text{CE} = -\frac{1}{T} \sum_{t=1}^{T} \sum_{c=1}^{4} s_{t}^{c} \log(\hat{s}_{t}^{c}),
\end{equation}

where $s_{t}^{c}$ represents the ground truth probability of game being in state $c$ at time $t$, and $\hat{s}_{t}^{c}$ is the predicted probability. The overall loss is  $ \mathcal{L} =  \mathcal{L}_\textrm{ADE} + \lambda  \mathcal{L}_\textrm{CE}$, where $\lambda$ is a weighting factor set to $\lambda=4$ when classifying states. In the supplementary (\textit{suppl}), we detail the training procedure and the chosen hyperparameters.

We will study the importance of each part of the method by considering: \textit{Ours w/o CLS} (without utilizing state classification), \textit{Ours w/o $\mathbf{M_{unc}}$} (using the binary mask $\mathbf{M}$ instead of $\mathbf{M_{unc}}$ in the loss term), and \textit{Ours w/o SOC} (without employing SAB$_S$ nor state classification). Additionally, we depict combinations of these variations.

\section{Experimental Evaluation}
\label{sec:results}

We next present experimental results on trajectory completion and state classification, comparing our approach with competing methods.
For quantitative evaluation, we utilize the ADE metric in
Eq.~\eqref{eq:Loss} but considering the binary mask $\mathbf{M}$ instead of $\mathbf{M_{unc}}$. For trajectory forecasting, we use the Final Displacement Error (FDE) to measure the final prediction deviation. We also consider the Maximum Error (MaxErr) to capture the largest discrepancies:
\begin{equation}
\nonumber
\small
    \textrm{MaxErr} = \frac{1}{D} \sum_{n=1}^{N}  \max_{t \in \{1, \ldots, T\}} \left( \left\lVert \mathbf{x}^{n}_{t} - \hat{\mathbf{x}}^{n}_{t} \right\rVert_2 \cdot \mathbf{M}_{t}^{n} \right),
\end{equation}
where $D = \sum_{n=1}^{N} \mathbbm{1}\left(\sum_{t=1}^{T} \mathbf{M}_{t}^{n}\right)$, with $\mathbbm{1}(\cdot)$ as the unit step function. For state classification, being $\mathbb{I(\cdot)}$ the indicator function, Accuracy (Acc) is computed as:

\begin{equation}
\nonumber
\small
    \textrm{Acc} = \frac{1}{T} \sum_{t=1}^{T} \mathbb{I} \left[ \argmax_{c} (s_{t}^{c}) = \argmax_{c} (\hat{s}_{t}^{c}) \right].
\end{equation}

\subsection{Datasets}
\label{subsec:dataset}

\noindent \textbf{Soccer:} This dataset comprises real soccer match data from LaLiga's 2022-2023 season, including 283 matches. The matches are split into sequences of $T=60$ frames, representing 9.6 seconds sampled at 6.25 Hz. Each frame contains 23 observations $(x,y)$ for each one of the agents (22 players and the ball). The \textit{agent type} is known in this dataset. Goalkeepers may contain NaNs if they are not visible. To ensure consistency with prior research, the agent order is standardized. The dataset is split into 82,954 / 7,500 / 6,258 sequences for training, validation and testing, respectively, with each split using different matches. For the state classification task, the dataset is complemented with one state label per frame, considering the states {\em pass}, {\em possession}, {\em uncontrolled} and {\em out-of-play}.

\noindent \textbf{Basketball-VU:} This dataset consists of basketball player tracking data provided by STATS SportVU from the 2016 NBA season. To evaluate our model on player forecasting, we use the same splits as in \cite{monti2021dag}. Each sequence consists of 50 timesteps representing 10 seconds sampled at 5Hz, where each one contains the $(x,y)$ observations for 10 players and the ball.

\noindent \textbf{Basketball-TIP:} We also consider another basketball dataset from the 2012 NBA season. Following the same splits as~\cite{zhan2018generating}, Xu {\em et al.}~\cite{xu2023uncovering} pre-processed it allowing to evaluate in both player imputation and forecasting tasks, renaming it as Basketball-TIP. This dataset employs two distinct strategies to simulate the appearance and disappearance of players: the ``circle mode'' and the ``camera mode''. Each sequence consists on 50 frames representing 8 seconds sampled at 6.25Hz, each one containing the $(x,y)$ observations for 10 players and the ball.

\noindent \textbf{ETH-UCY:} For completeness, we conducted an experiment using the ETH-UCY pedestrian dataset~\cite{pellegrini2009you, lerner2007crowds}. Our approach performs comparably to deterministic SOTA architectures~\cite{saadatnejad2023social, xu2023eqmotion}, achieving a 4.3\% improvement in ADE on the ETH subset. Detailed information and results are available in the \textit{suppl}.

\subsection{Player Forecasting and Imputation}
\label{subsec:playerforec}

First, we assess our model's effectiveness in  \textbf{(i) soccer player forecasting and imputation}. The predicted players, referred to as agents of interest $P$, are predicted using all future visible observations of conditioning agents, like the ball and/or an opponent team. In the forecasting task, the model observes 20 timesteps (3.2s) and predicts the next 40 timesteps (6.4s) of $P$. The imputation task is similar but with the final location of each player of interest set as visible. As in previous studies~\cite{yeh2019diverse}, goalkeepers are excluded from this analysis.

In the forecasting task, we compare against the following implemented baselines: \textit{Velocity} extrapolation, projecting agent predictions linearly based on observed velocity; \textit{RNN} encoder with LSTM, using shared weights, and MLP decoder for prediction~\cite{becker2018red}; \textit{GRNN} as the non-variational version of GVRNN~\cite{yeh2019diverse}; \textit{GRNN+Att} which is the previous baseline but using GAT instead of GNNs; \textit{Transformer} which mirrors our pipeline but uses SAB to perform attention across all timesteps of all agents simultaneously~\cite{alcorn2021baller2vec++}, as opposed to decoupling attention in SAB$_T$ and SAB$_S$; and \textit{Ours w/o SOC}. Further details of these implementations can be found in the \textit{suppl}. It is important to note that \textit{Velocity}, \textit{RNN}, and \textit{Ours w/o SOC} operate independently for each agent, making the agents ordering irrelevant and preventing them from utilizing any social conditioning.

\begin{table*}[t]
    \centering
    \resizebox{12 cm}{!} {
    \begin{tabular}{cl*{11}{@{\hspace{6mm}}c}}
    \toprule
    \multirow{3}{*}{\begin{tabular}[c]{@{}c@{}}Predict $P$\\ (Condition)\end{tabular}}
    & & \multicolumn{8}{c}{Forecasting} & \multicolumn{2}{c}{Imputation} \\
    \cmidrule(rr){3-10} \cmidrule(rr){11-12} 
    &
    & \text{Velocity}
    & \text{RNN}
    & \text{Ours w/o SOC}
    & \text{GRNN}
    & \text{GRNN+Att}
    & \text{Transformer}
    & \text{Ours w/o CLS}
    & \text{Ours}
    & \text{Ours w/o CLS}
    & \text{Ours} \\
    \cmidrule{2-12}
     & Social & & & & \checkmark & \checkmark & \checkmark & \checkmark & \checkmark & \checkmark & \checkmark\\
    \midrule
    \multirow{5}{*}{\begin{tabular}[c]{@{}c@{}}Players\\ (Ball)\end{tabular}} & ADE$_{P}$ $\downarrow$ & 5.96 & 4.36 & 4.08 & 4.02 & 3.67 & 2.66 & 2.53 & \textbf{2.42} & \textbf{1.14} & 1.15\\
                              & MaxErr$_{P}$ $\downarrow$ & 13.49 & 8.95 & 8.60 & 7.43 & 7.02 & 5.28 & 5.12 & \textbf{4.97} & \textbf{2.21} & 2.22\\
                              & FDE$_{P}$ $\downarrow$ & 13.33 & 8.59 & 8.25 & 6.85 & 6.49 & 4.78 & 4.65 & \textbf{4.50} & - & -\\
                              & Acc (\%) $\uparrow$ & - & - & - & - & - & - & - & 87.35 & - & 89.00\\
    \midrule
    \multirow{5}{*}{\begin{tabular}[c]{@{}c@{}}Offense\\ (Defense+Ball)\end{tabular}} & ADE$_{P}$ $\downarrow$ & 5.76 & 4.23 & 3.96 & 3.76 & 3.30 & 2.26 & 2.10 & \textbf{2.06} & \textbf{0.99} & 1.02\\
                              & MaxErr$_{P}$ $\downarrow$ & 13.04 & 8.72 & 8.39 & 6.84 & 6.31 & 4.45 & 4.27 & \textbf{4.21} & \textbf{1.92} & 1.97\\
                              & FDE$_{P}$ $\downarrow$ & 12.89 & 8.39 & 8.07 & 6.32 & 5.80 & 3.96 & 3.82 & \textbf{3.77} & - & -\\
                              & Acc (\%) $\uparrow$ & - & - & - & - & - & - & - & 88.91 & - & 89.69\\
    \midrule
    \multirow{5}{*}{\begin{tabular}[c]{@{}c@{}}Defense\\ (Offense+Ball)\end{tabular}} & ADE$_{P}$ $\downarrow$ & 6.16 & 4.49 & 4.20 & 3.47 & 3.22 & 2.14 & 2.01 & \textbf{1.98} & \textbf{1.03} & 1.04\\
                              & MaxErr$_{P}$ $\downarrow$ & 13.94 & 9.18 & 8.81 & 6.29 & 5.98 & 4.17 & 4.04 & \textbf{3.98} & \textbf{1.99} & 2.00\\
                              & FDE$_{P}$ $\downarrow$ & 13.78 & 8.79 & 8.44 & 5.69 & 5.36 & 3.63 & 3.55 & \textbf{3.49} & - & -\\
                              & Acc (\%) $\uparrow$ & - & - & - & - & - & - & - & 89.92 & - & 90.47\\
    \midrule
    \end{tabular}}
    \vspace{0.1cm}
    \caption{\textbf{Evaluation in (i) soccer player forecasting and imputation.} Predictions are generated with a time horizon of 6.4s using a prior of 3.2s. $P$ denotes agents of interest. For the imputation task, the last observation of each agent is visible. All metrics, except Acc, are in meters. 
    }
    \label{tab:playerforecasting}
\end{table*}
\begin{figure*}[t!]
  \centering
  \resizebox{10.2 cm}{!} {
  \includegraphics[trim={0 0.25cm 0 0},clip]{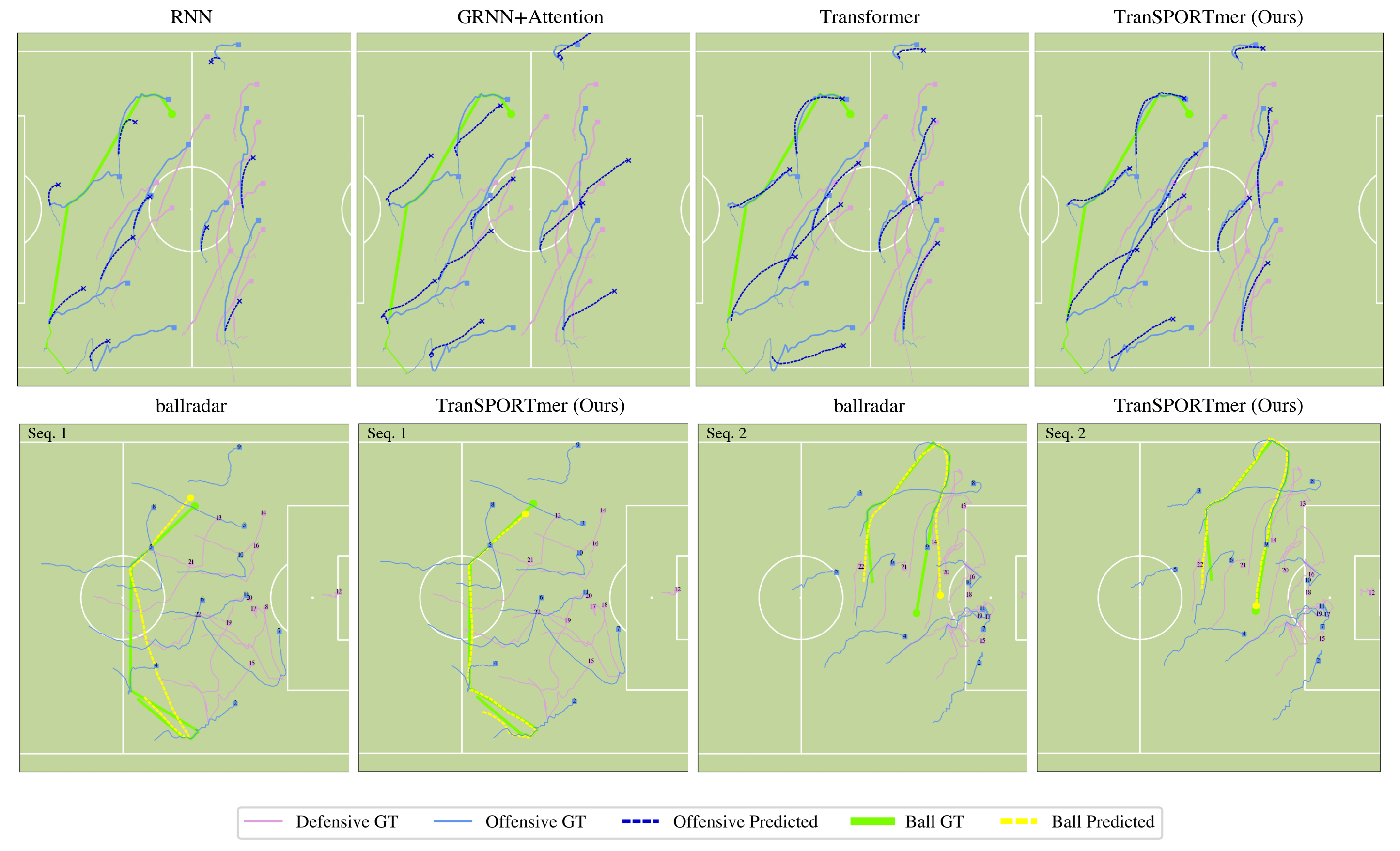}}
\caption{\textbf{Qualitative evaluation in soccer player forecasting and ball inference}. \textbf{Top:} 
Offensive player trajectory forecasting with a time horizon of 6.4s using a prior of 3.2s.
\textbf{Bottom:} Ball inference through the full 9.6s sequence.}
  \label{fig:FigQual1}
\end{figure*}

\Cref{tab:playerforecasting} shows the results of (i) soccer player forecasting and imputation with conditioning agents indicated in parentheses. As expected, socially aware architectures exhibit superior performance in all metrics, particularly when the number of conditioning agents is increased.  Results for \textit{Ours w/o SOC} underscore the clear significance of SAB$_S$. \textit{Transformer} achieves slightly inferior results compared to \textit{Ours w/o CLS}, likely due to its flattened attention mechanism, which may cause confusion with the higher number of non-correlated observations. Additionally, \textit{Transformer} is approximately four times slower at inference time (340 vs. 88 milliseconds). Furthermore, \methodname~(\textit{Ours}) outperforms \textit{Ours w/o CLS} in forecasting, but in the imputation task, \textit{Ours w/o CLS} achieves slightly better results, possibly due to sub-optimal $\lambda$ compared to forecasting. \Cref{fig:FigQual1}-top shows an example on the task of forecasting offensive players (second task-row in~\Cref{tab:playerforecasting}). The \textit{RNN} baseline tends to generate shorter predicted trajectories, emphasizing the need for social interactions to enhance performance. The \textit{GRNN+Att} baseline exhibits improved performance with conditioning in long-term predictions. However, \methodname~outperforms these baselines, yielding more realistic results aligned with ground truth positions (see video of our results in the \textit{suppl}).

\begin{figure}[t]
  \centering
  \begin{subfigure}{0.54\textwidth}
    \centering
    \includegraphics[width=\linewidth,valign=c]{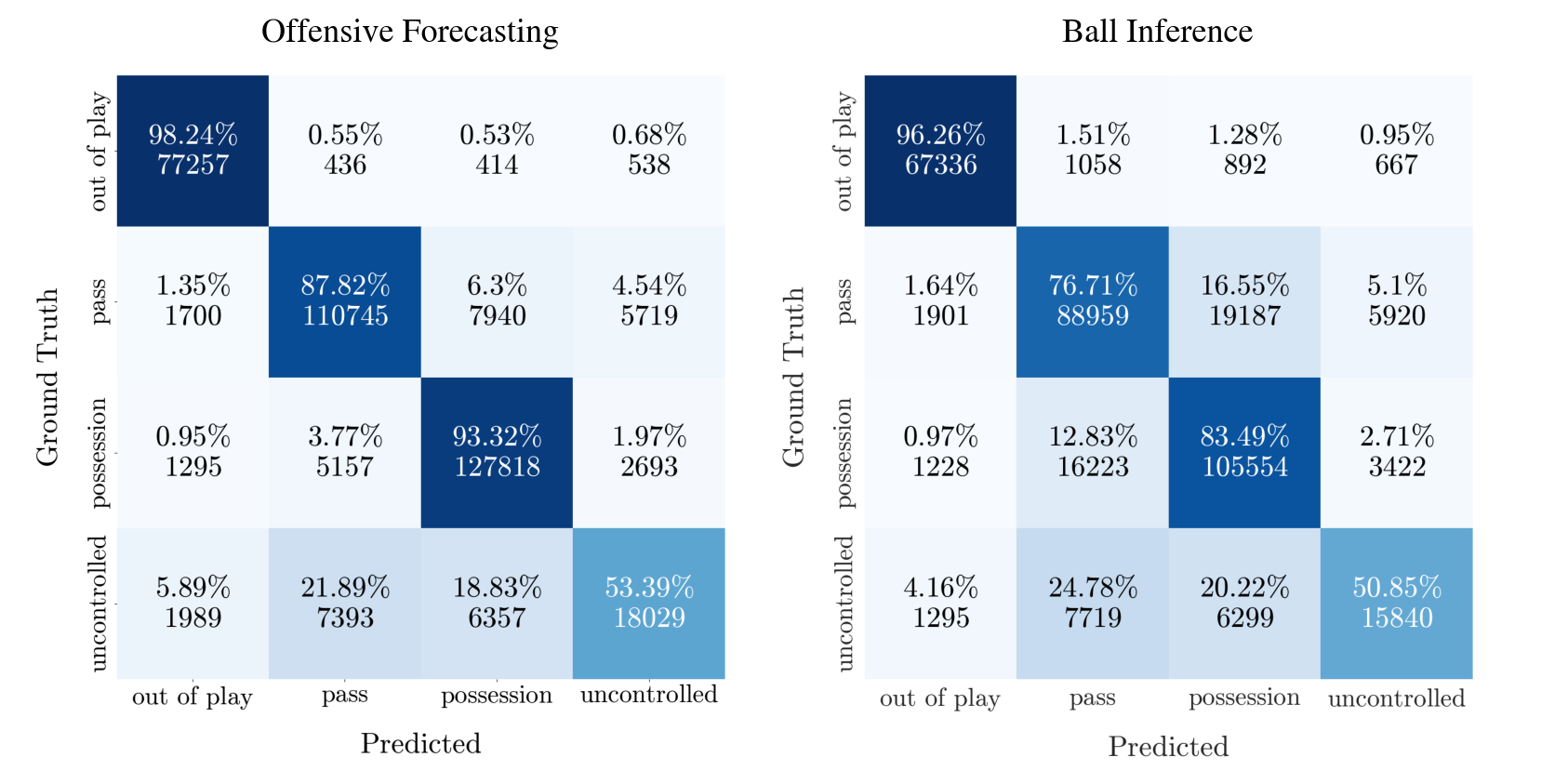}
  \end{subfigure}
  \hfill
  \begin{subfigure}{0.45\textwidth}
    \centering
    \includegraphics[width=\linewidth,valign=c]{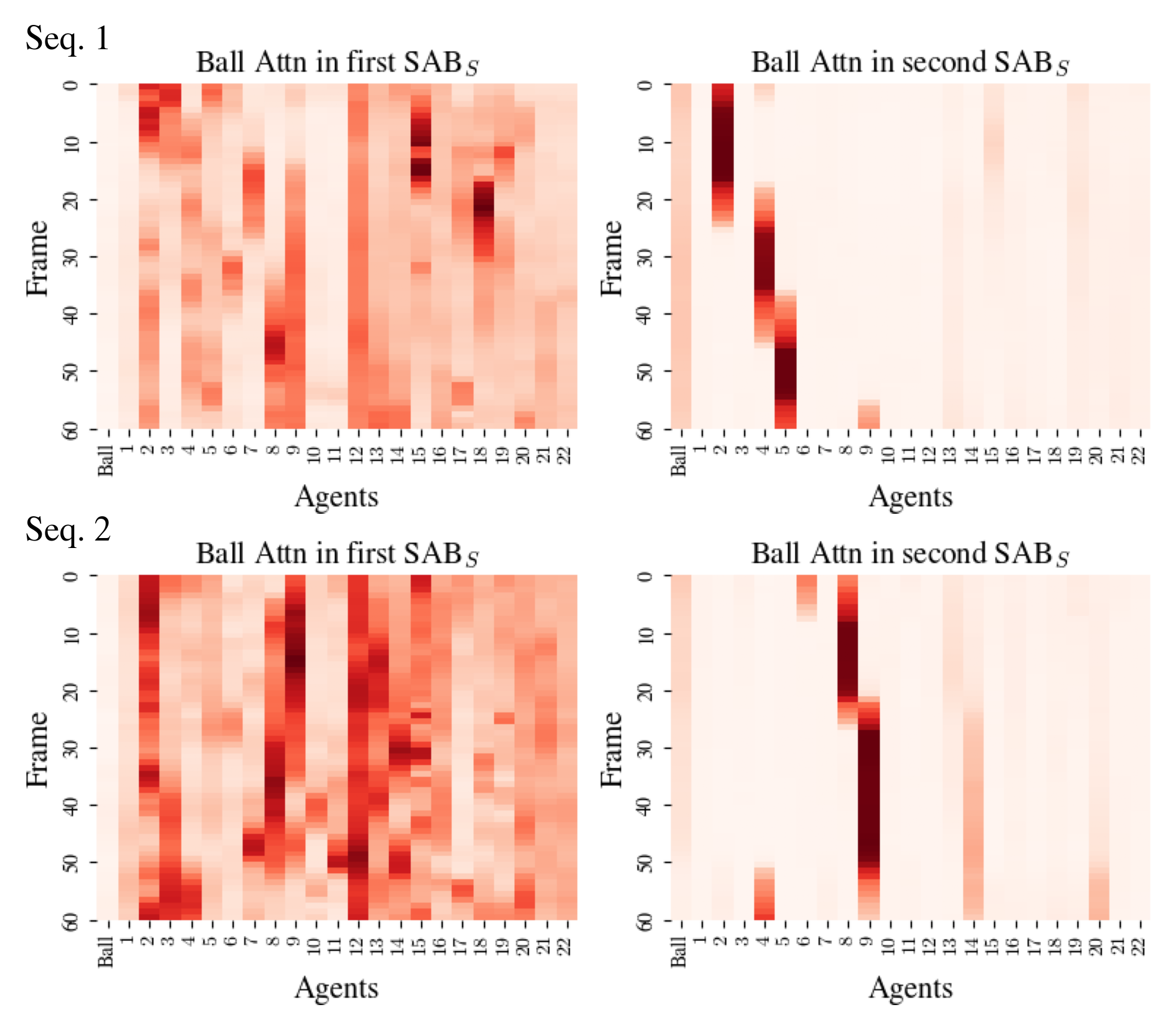}
  \end{subfigure}
  \caption{\textbf{Left: Confusion matrix in state classification.} Offensive player trajectory forecasting (left) and ball inference (right). \textbf{Right: Attention maps for the ball.} Visualization of attention maps in each social SAB$_S$ across agents and time for the sequences \#1 and \#2 in Fig.~\ref{fig:FigQual1}-bottom (animations in \textit{suppl} video).}
  \label{fig:confusion-attention}
\end{figure}

~\Cref{tab:playerforecasting} also reports the accuracy of state classification while addressing trajectory prediction and imputation tasks. Achieving approximately 90\%, these results demonstrate the robust and consistent classification power of our model, primarily attributed to the ball's visibility in all tasks. The confusion matrix in Fig.~\ref{fig:confusion-attention}-left-left specifically illustrates the state classification while forecasting offensive players, achieving an overall accuracy of 88.91\%. It is worth noting that the \textit{uncontrolled} class exhibits less accurate predictions due to its challenging subjective nature in annotations and an imbalance compared to the other classes.

\begin{table*}[t!]
  \centering
    \resizebox{12 cm}{!} {
  \begin{tabular}{lccccccccc}
    \toprule
      \multirow{2}{*}{Predict $P$}
      &  \hspace{6pt}STGAT~\cite{huang2019stgat}
      &  \hspace{6pt}Social-Ways~\cite{amirian2019social}
      &  \hspace{6pt}GVRNN~\cite{yeh2019diverse}
      &  \hspace{6pt}GMAT~\cite{zhan2018generating}
      &  \hspace{6pt}AC-VRNN~\cite{bertugli2021ac}
      &  \hspace{6pt}DAG-Net~\cite{monti2021dag}
      &  \hspace{6pt}U-MAT~\cite{fassmeyer2022semi}
      &  \hspace{6pt}S-PatteRNN~\cite{navarro2022social}
      &  \multirow{2}{*}{\hspace{6pt}Ours w/o CLS}\\
    \cmidrule(lr){2-9}
    & ICCV'19 & CVPRW'19 & CVPR'19 & ICLR'19 & CVIU'21 & ICPR'21 & NeurIPS'22 & IROS'22 \\
    \midrule
    Players & - & - & - & - & - & 8.55/12.37 & - & 8.13/12.34 & \textbf{7.75}/\textbf{11.65}  \\
    Offense  & 9.94/15.80 & 9.91/15.19 & 9.73/15.89 & 9.47/16.98 & 9.32/14.91  & \textbf{8.98}/14.08 & 9.01/\textbf{13.28} &  - & 9.19/14.24\\
    Defense & 7.26/11.28 & 7.31/10.21 & 7.29/10.62 & 7.05/10.56 & 7.01/10.16 & 6.87/9.76 & 6.88/\textbf{9.04} &  - & \textbf{6.31}/\textbf{9.04}\\
    \bottomrule
  \end{tabular}}
  \vspace{0.2cm}
  \caption{\textbf{Evaluation in (ii) basketball player forecasting using Basketball-VU dataset (ADE$_P$/FDE$_P$).} Predictions have a time horizon of 8s using a prior of 2s. Results are extracted from the original works, and no agent future condition is considered in this task. $P$ denotes agents of interest. All metrics are in feet. 
  }
  \label{tab:playerforecasting_basketball}
\end{table*}
\begin{table}[t!]
  \centering
  \scriptsize	
  \resizebox{12 cm}{!}{
  \begin{tabular}{@{} ll @{\hspace{4pt}} c @{\hspace{2pt}} c @{\hspace{2pt}} c @{\hspace{2pt}} c @{\hspace{2pt}} c @{\hspace{2pt}} c @{\hspace{2pt}} c @{\hspace{2pt}} c @{\hspace{2pt}} c @{} c @{\hspace{2pt}} c @{} c}
    \toprule
    \multirow{2}{*}{Model}
    &
    & \multicolumn{2}{c}{$r = 3$ft} 
    & \multicolumn{2}{c}{$r = 5$ft}
    & \multicolumn{2}{c}{$r = 7$ft}
    & \multicolumn{2}{c}{$\theta = 10$º} 
    & \multicolumn{2}{c}{$\theta = 20$º}
    & \multicolumn{2}{c}{$\theta = 30$º}\\
    && \textbf{I-}ADE & \textbf{P-}ADE & \textbf{I-}ADE & \textbf{P-}ADE & \textbf{I-}ADE & \textbf{P-}ADE & \textbf{I-}ADE & \textbf{P-}ADE & \textbf{I-}ADE & \textbf{P-}ADE & \textbf{I-}ADE & \textbf{P-}ADE \\
    \midrule
    Mean && 9.07 (10.36) & - & 9.53 (9.44) & - & 9.51 (9.21) & - & 8.83 (8.56) & - & 8.64 (8.73) & - & 8.47 (8.92) & - \\
    Median && 9.32 (10.55) & - & 9.82 (9.64) & - & 9.81 (9.44) & - & 9.16 (8.84) & - & 8.96 (9.02) & - & 8.75 (9.21) & - \\
    GMAT \cite{zhan2018generating} & ICLR'19 & 7.36 & - & 6.89 & - & 6.73 & - & 6.42 & - & 5.99 & - & 6.01 & - \\
    NAOMI \cite{liu2019naomi} & NeurIPS'19 & 7.68 & - & 7.08 & - & 7.04 & - & 6.33 & - & 6.11 & - & 5.91 & - \\
    LSTM \cite{hochreiter1997long} & NeurComp& 7.33 & 20.07 & 6.73 & 14.91 & 6.51 & 10.07 & 6.28 & 9.34 & 6.01 & 7.52 & 5.67 & 6.10 \\
    VRNN \cite{chung2015recurrent} & NeurIPS'15 & 7.43 & 12.26 & 6.90 & 11.38 & 6.68 & 10.07 & 6.38 & 8.49 & 6.09 & 7.47 & 5.92 & 7.36 \\
    INAM \cite{qi2020imitative} \hspace{10pt}& CVPR'20 & 7.35 & 8.93 & 6.93 & 8.24 & 6.80 & 7.68 & 6.50 & 7.32 & 6.13 & 7.10 & 5.92 & 6.96 \\
    GC-VRNN \cite{xu2023uncovering}& CVPR'23 & 7.03 & 8.93 & 6.93 & 8.24 & 6.80 & 7.68 & 5.86 & 6.29 & 5.56 & 4.74 & 5.39 & 4.28 \\
    \midrule
    Our w/o CLS/$\mathbf{M_{unc}}$ && (5.32) & (5.91) & (4.71) & (5.56) & (4.16) & (4.91) & (3.60) & (\textbf{4.77}) & (3.29) & (4.13) & (\textbf{3.08}) & (\textbf{3.60}) \\
    Our w/o CLS && (\textbf{5.24}) & (\textbf{5.89}) & (\textbf{4.48}) & (\textbf{5.29}) & (\textbf{4.14}) & (\textbf{4.90}) & (\textbf{3.59}) & (4.78) & (\textbf{3.26}) & (\textbf{4.09}) & (\textbf{3.08}) & (\textbf{3.60}) \\
    \bottomrule
  \end{tabular}}
  \vspace{0.2cm}
  \caption{\textbf{Evaluation in (iii) basketball player unified imputation and forecasting using the Basketball-TIP dataset~\cite{xu2023uncovering}.} The imputation task is performed over 6.4 seconds, and forecasting over 1.6 seconds. All metrics are in feet. Our implementation results are presented in parentheses.}
  \label{tab:basketball_imput_forec}
\end{table}

Next, we evaluate the effectiveness of our model in \textbf{(ii) basketball player forecasting} using the Basketball-VU dataset. The task at hand consists of observing 10 time-steps (2s) and predicting the following 40 (8s) of players without future conditioning agents (refer to \textit{suppl} for additional conditioning-based experiments). We compare against the state-of-the-art results already published in previous works, as shown in Table~\ref{tab:playerforecasting_basketball}. Our model is trained to predict both offensive and defensive players simultaneously. Other baselines, like DAG-Net~\cite{monti2021dag}, need separate training to achieve better results. The ADE$_P$ and FDE$_P$ metrics depicted in the table demonstrate that our method outperforms in predicting trajectories for all players and defense, using only one model trained with the same weights. In both Soccer and Basketball-VU datasets, it can be seen that in general, forecasting offensive players is more challenging than defensive ones.

Additionally, we assess our model's capability in \textbf{(iii) basketball player unified imputation and forecasting} tasks using the Basketball-TIP dataset. This task involves observing the initial 40 timesteps (6.4s), imputing agents outside the circle/camera view, and forecasting their locations during the subsequent 10 frames (1.6s). In ``circle mode'', three radii $r\in\{3,5,7\}$ ft are considered, centered on the ball location. In ``camera mode'', a fixed field of view (FOV) tracks the ball from the center of the pitch, with three possible angles $\theta \in\{10,20,30\}^{\circ}$. Following Xu {\em et al.}~\cite{xu2023uncovering}, we predict players who have at least one observation in the initial 40 timesteps, potentially varying numbers of agents across sequences. Our method incorporates the additional NaN-mask to exclude non-interest agents within each sequence. We add our results in Table~\ref{tab:basketball_imput_forec}, showing a clear effectiveness of our method against the SOTA approaches in all six scenarios. \textbf{I-}ADE denotes the error in the initial 40 timesteps (imputation error) and \textbf{P-}ADE signifies the error in the final 10 timesteps (forecasting error). Our method performs notably well in imputation tasks compared to GC-VRNN~\cite{xu2023uncovering}, due to its unidirectional recurrent nature, which affects forecasting reliability based on imputed data. Refer to the \textit{suppl} for detailed information and figures.

\subsection{Ball Imputation and Inference}
We evaluate \textbf{(iv) soccer ball imputation and inference} tasks. The inference task involves predicting all observations of the ball, masking 100\% of them. The imputation task involves predicting a lower percentage of the ball observations while setting the others as visible. Players' observations serve as conditioning agents in all tasks. We benchmark against the state-of-the-art method \textit{ballradar} \cite{kim2023ball}, which employs a hierarchical approach involving possessor classification followed by ball trajectory regression. Additionally, we compare against its non-hierarchical version, \textit{ballradar w/o POS}, which performs ball regression without possessor classification. Due to the requirements of \textit{ballradar}, our dataset is augmented with ground-truth possessor information, player's velocities and goalkeeper locations using our method (further details can be found in the \textit{suppl}).

Table~\ref{tab:BallInf1} presents a comparative analysis of ball trajectory imputation and inference. Our methods consistently outperform the state-of-the-art, achieving over a 25\% improvement in ADE for trajectory inference. Qualitative differences in two test samples are illustrated in Fig.~\ref{fig:FigQual1}-bottom. For imputation, we showcase results by masking 80\% and 90\% of total ball observations for each sequence, highlighting the superior performance of our method.  Notably, employing state classification in \methodname~helps achieve generally better results, showcasing its holistic nature. However, \textit{Ours w/o CLS} surpasses \textit{ballradar} without requiring additional data beyond player $(x,y)$ locations.

In terms of state classification accuracy (see \Cref{tab:BallInf1}), there is an anticipated decline compared to the soccer player trajectory forecasting and imputation section (see \Cref{tab:playerforecasting}), likely attributed to the non-visibility of the ball, our target. Surprisingly, the method still achieves an accuracy of 80.84\% in state classification showcasing that the game states can be inferred using only the movement of players (refer to Fig.~\ref{fig:confusion-attention}-left-right for the detailed confusion matrix). Figure~\ref{fig:confusion-attention}-right shows the attention maps generated by the SAB$_S$ for the ball across all agents and timesteps in the two examples of Fig.~\ref{fig:FigQual1}-bottom. Computed by averaging contributions from each head, these maps reveal the model's awareness. In the first SAB$_S$, a broad, general awareness of other agents is observed, resembling a coarse social perspective. The second SAB$_S$ focuses attention on the possessor player or the anticipated recipient of the ball in the event of a pass. This highlights the coarse-to-fine nature of the two encoders in our model. Refer to the \textit{suppl} for additional ablation study regarding the coarsening-to-fine.

\begin{table}[t!]
  \centering
  \resizebox{12 cm}{!} {
  \begin{tabular}{lc<{\hspace{8pt}}*{5}{c<{\hspace{8pt}}c<{\hspace{8pt}}c<{\hspace{10pt}}}}
    \toprule
    & ballradar w/o POS
    & \multicolumn{3}{c}{ballradar} (KDD'23)
    & \multicolumn{3}{c}{Ours w/o CLS/$\mathbf{M_{unc}}$} 
    & \multicolumn{3}{c}{Ours w/o CLS} 
    & \multicolumn{3}{c}{Ours w/o $\mathbf{M_{unc}}$} 
    & \multicolumn{3}{c}{Ours}  \\
    \cmidrule(rr){2-2} \cmidrule(rr){3-5} \cmidrule(rr){6-8} \cmidrule(rr){9-11} \cmidrule(rr){12-14}  \cmidrule(rr){15-17}
    Mask & 100\% & 80\% & 90\% & 100\% & 80\% & 90\% & 100\% & 80\% & 90\% & 100\% & 80\% & 90\% & 100\% & 80\% & 90\% & 100\% \\
    \midrule
     ADE $\downarrow$ & 5.43 & 0.97 & 1.51 & 3.89 & 0.88 & 1.18 & 2.89 & 0.88 & 1.12 & 2.89 &\textbf{0.80} & 1.23 & \textbf{2.71} & 0.84 & \textbf{1.09} & \textbf{2.71} \\
     MaxErr $\downarrow$ & 10.98 & 3.73 & 5.16 & 8.79 & 3.47 & 4.59 & 7.78 &  3.44 & 4.48 & 7.78 & 3.25 & 4.48 & \textbf{7.39} & \textbf{3.24} & \textbf{4.39} & \textbf{7.39} \\
     Acc (\%) $\uparrow$ & - & - & - & -& - & - & - & - & - & - & 85.51 & 83.38 & \textbf{80.84} & \textbf{85.59} & \textbf{83.55} & \textbf{80.84}\\
    \bottomrule
  \end{tabular}}
  \vspace{0.2cm}
  \caption{\textbf{Evaluation in (iv) soccer ball imputation and inference.} Predictions are generated through the full 9.6s sequence. All metrics, except Acc, are in meters.}
  \label{tab:BallInf1}
\end{table} 

In both Tables~\ref{tab:basketball_imput_forec} and \ref{tab:BallInf1}, we include an ablation study regarding the usage of $\mathbf{M_{unc}}$ in the loss term, which generally leads to improved results. For the first neighbors, the recorded values are $w_1 \in [0.7, 0.85]$, and for the second ones $w_2 \in [0.15, 0.3]$, reflecting the expected level of uncertainty.

\section{Conclusions}
\label{sec:conclusions}
In this paper, we introduced \methodname, a holistic approach capable of handling multiple tasks (forecasting, imputation, inference, and state classification) for trajectory understanding in multi-agent sports scenarios. Unlike state-of-the-art methods, \methodname~can address all tasks using our approach, eliminating the need for task-specific models. Our evaluation on soccer and basketball datasets shows competitive performance across tasks. Notably, our approach excels in player forecasting, player imputation, ball imputation and inference tasks, while combined with state classification tasks allows to improve the results. Additionally, the learnable mask models uncertainty in neighboring hidden values, further enhancing outcomes. We believe this has the potential to pave the way for a deeper understanding of the semantic aspects of sports games.

\noindent \textbf{Acknowledgment.} This work has been supported by the project GRAVATAR PID2023-151184OB-I00 funded by MCIU/AEI/10.13039/501100011033 and by ERDF, UE and by the Government of Catalonia under 2023 DI 00058.


%
%
\bibliographystyle{splncs04}
\bibliography{main}

\begin{thebibliography}{10}
\providecommand{\url}[1]{\texttt{#1}}
\providecommand{\urlprefix}{URL }
\providecommand{\doi}[1]{https://doi.org/#1}

\bibitem{aksan2021spatio}
Aksan, E., Kaufmann, M., Cao, P., Hilliges, O.: A spatio-temporal transformer for 3d human motion prediction. In: 2021 International Conference on 3D Vision (3DV). pp. 565--574. IEEE (2021)

\bibitem{alahi2016social}
Alahi, A., Goel, K., Ramanathan, V., Robicquet, A., Fei-Fei, L., Savarese, S.: Social lstm: Human trajectory prediction in crowded spaces. In: Proceedings of the IEEE conference on computer vision and pattern recognition. pp. 961--971 (2016)

\bibitem{alcorn2021baller2vec++}
Alcorn, M.A., Nguyen, A.: baller2vec++: A look-ahead multi-entity transformer for modeling coordinated agents. arXiv preprint arXiv:2104.11980  (2021)

\bibitem{alcorn2021baller2vec}
Alcorn, M.A., Nguyen, A.: baller2vec: A multi-entity transformer for multi-agent spatiotemporal modeling. arXiv preprint arXiv:2102.03291  (2021)

\bibitem{amirian2019social}
Amirian, J., Hayet, J.B., Pettr{\'e}, J.: Social ways: Learning multi-modal distributions of pedestrian trajectories with gans. In: Proceedings of the IEEE/CVF Conference on Computer Vision and Pattern Recognition Workshops. pp.~0--0 (2019)

\bibitem{amirli2022prediction}
Amirli, A., Alemdar, H.: Prediction of the ball location on the 2d plane in football using optical tracking data. Academic Platform Journal of Engineering and Smart Systems  \textbf{10}(1), ~1--8 (2022)

\bibitem{battaglia2018relational}
Battaglia, P.W., Hamrick, J.B., Bapst, V., Sanchez-Gonzalez, A., Zambaldi, V., Malinowski, M., Tacchetti, A., Raposo, D., Santoro, A., Faulkner, R., et~al.: Relational inductive biases, deep learning, and graph networks. arXiv preprint arXiv:1806.01261  (2018)

\bibitem{becker2018red}
Becker, S., Hug, R., Hubner, W., Arens, M.: Red: A simple but effective baseline predictor for the trajnet benchmark. In: Proceedings of the European Conference on Computer Vision (ECCV) Workshops. pp.~0--0 (2018)

\bibitem{bertugli2021ac}
Bertugli, A., Calderara, S., Coscia, P., Ballan, L., Cucchiara, R.: Ac-vrnn: Attentive conditional-vrnn for multi-future trajectory prediction. Computer Vision and Image Understanding  \textbf{210},  103245 (2021)

\bibitem{brito2020association}
Brito~Souza, D., L{\'o}pez-Del~Campo, R., Blanco-Pita, H., Resta, R., Del~Coso, J.: Association of match running performance with and without ball possession to football performance. International Journal of Performance Analysis in Sport  \textbf{20}(3),  483--494 (2020)

\bibitem{cai2020learning}
Cai, Y., Huang, L., Wang, Y., Cham, T.J., Cai, J., Yuan, J., Liu, J., Yang, X., Zhu, Y., Shen, X., et~al.: Learning progressive joint propagation for human motion prediction. In: Computer Vision--ECCV 2020: 16th European Conference, Glasgow, UK, August 23--28, 2020, Proceedings, Part VII 16. pp. 226--242. Springer (2020)

\bibitem{cao2018brits}
Cao, W., Wang, D., Li, J., Zhou, H., Li, L., Li, Y.: Brits: Bidirectional recurrent imputation for time series. Advances in neural information processing systems  \textbf{31} (2018)

\bibitem{capellera2024footbots}
Capellera, G., Ferraz, L., Rubio, A., Agudo, A., Moreno-Noguer, F.: Footbots: A transformer-based architecture for motion prediction in soccer. In: 2024 IEEE International Conference on Image Processing (ICIP). pp. 2313--2319. IEEE (2024)

\bibitem{chung2015recurrent}
Chung, J., Kastner, K., Dinh, L., Goel, K., Courville, A.C., Bengio, Y.: A recurrent latent variable model for sequential data. Advances in neural information processing systems  \textbf{28} (2015)

\bibitem{decroos2019actions}
Decroos, T., Bransen, L., Van~Haaren, J., Davis, J.: Actions speak louder than goals: Valuing player actions in soccer. In: Proceedings of the 25th ACM SIGKDD international conference on knowledge discovery \& data mining. pp. 1851--1861 (2019)

\bibitem{decroos2018automatic}
Decroos, T., Van~Haaren, J., Davis, J.: Automatic discovery of tactics in spatio-temporal soccer match data. In: Proceedings of the 24th acm sigkdd international conference on knowledge discovery \& data mining. pp. 223--232 (2018)

\bibitem{devlin2018bert}
Devlin, J., Chang, M.W., Lee, K., Toutanova, K.: Bert: Pre-training of deep bidirectional transformers for language understanding. arXiv preprint arXiv:1810.04805  (2018)

\bibitem{ding2020graph}
Ding, D., Huang, H.H.: A graph attention based approach for trajectory prediction in multi-agent sports games. arXiv preprint arXiv:2012.10531  (2020)

\bibitem{dosovitskiy2020image}
Dosovitskiy, A., Beyer, L., Kolesnikov, A., Weissenborn, D., Zhai, X., Unterthiner, T., Dehghani, M., Minderer, M., Heigold, G., Gelly, S., et~al.: An image is worth 16x16 words: Transformers for image recognition at scale. arXiv preprint arXiv:2010.11929  (2020)

\bibitem{everett2023inferring}
Everett, G., Beal, R.J., Matthews, T., Early, J., Norman, T.J., Ramchurn, S.D.: Inferring player location in sports matches: Multi-agent spatial imputation from limited observations. arXiv preprint arXiv:2302.06569  (2023)

\bibitem{fassmeyer2021toward}
Fassmeyer, D., Anzer, G., Bauer, P., Brefeld, U.: Toward automatically labeling situations in soccer. Frontiers in Sports and Active Living  \textbf{3},  725431 (2021)

\bibitem{fassmeyer2022semi}
Fassmeyer, D., Fassmeyer, P., Brefeld, U.: Semi-supervised generative models for multiagent trajectories. Advances in Neural Information Processing Systems  \textbf{35},  37267--37281 (2022)

\bibitem{felsen2018will}
Felsen, P., Lucey, P., Ganguly, S.: Where will they go? predicting fine-grained adversarial multi-agent motion using conditional variational autoencoders. In: Proceedings of the European conference on computer vision (ECCV). pp. 732--747 (2018)

\bibitem{fragkiadaki2015recurrent}
Fragkiadaki, K., Levine, S., Felsen, P., Malik, J.: Recurrent network models for human dynamics. In: Proceedings of the IEEE international conference on computer vision. pp. 4346--4354 (2015)

\bibitem{girgis2021latent}
Girgis, R., Golemo, F., Codevilla, F., Weiss, M., D'Souza, J.A., Kahou, S.E., Heide, F., Pal, C.: Latent variable sequential set transformers for joint multi-agent motion prediction. arXiv preprint arXiv:2104.00563  (2021)

\bibitem{giuliari2021transformer}
Giuliari, F., Hasan, I., Cristani, M., Galasso, F.: Transformer networks for trajectory forecasting. In: 2020 25th international conference on pattern recognition (ICPR). pp. 10335--10342. IEEE (2021)

\bibitem{gu2017non}
Gu, J., Bradbury, J., Xiong, C., Li, V.O., Socher, R.: Non-autoregressive neural machine translation. arXiv preprint arXiv:1711.02281  (2017)

\bibitem{guo2023back}
Guo, W., Du, Y., Shen, X., Lepetit, V., Alameda-Pineda, X., Moreno-Noguer, F.: Back to mlp: A simple baseline for human motion prediction. In: Proceedings of the IEEE/CVF Winter Conference on Applications of Computer Vision. pp. 4809--4819 (2023)

\bibitem{gupta2018social}
Gupta, A., Johnson, J., Fei-Fei, L., Savarese, S., Alahi, A.: Social gan: Socially acceptable trajectories with generative adversarial networks. In: Proceedings of the IEEE conference on computer vision and pattern recognition. pp. 2255--2264 (2018)

\bibitem{hauri2021multi}
Hauri, S., Djuric, N., Radosavljevic, V., Vucetic, S.: Multi-modal trajectory prediction of nba players. In: Proceedings of the IEEE/CVF Winter Conference on Applications of Computer Vision. pp. 1640--1649 (2021)

\bibitem{hochreiter1997long}
Hochreiter, S., Schmidhuber, J.: Long short-term memory. Neural computation  \textbf{9}(8),  1735--1780 (1997)

\bibitem{honda2022pass}
Honda, Y., Kawakami, R., Yoshihashi, R., Kato, K., Naemura, T.: Pass receiver prediction in soccer using video and players' trajectories. In: Proceedings of the IEEE/CVF Conference on Computer Vision and Pattern Recognition. pp. 3503--3512 (2022)

\bibitem{hu2022entry}
Hu, B., Cham, T.J.: Entry-flipped transformer for inference and prediction of participant behavior. In: European Conference on Computer Vision. pp. 439--456. Springer (2022)

\bibitem{huang2019stgat}
Huang, Y., Bi, H., Li, Z., Mao, T., Wang, Z.: Stgat: Modeling spatial-temporal interactions for human trajectory prediction. In: Proceedings of the IEEE/CVF international conference on computer vision. pp. 6272--6281 (2019)

\bibitem{jain2016structural}
Jain, A., Zamir, A.R., Savarese, S., Saxena, A.: Structural-rnn: Deep learning on spatio-temporal graphs. In: Proceedings of the ieee conference on computer vision and pattern recognition. pp. 5308--5317 (2016)

\bibitem{kim2023ball}
Kim, H., Choi, H.J., Kim, C.J., Yoon, J., Ko, S.K.: Ball trajectory inference from multi-agent sports contexts using set transformer and hierarchical bi-lstm. arXiv preprint arXiv:2306.08206  (2023)

\bibitem{kong2020sound}
Kong, Q., Xu, Y., Wang, W., Plumbley, M.D.: Sound event detection of weakly labelled data with cnn-transformer and automatic threshold optimization. IEEE/ACM Transactions on Audio, Speech, and Language Processing  \textbf{28},  2450--2460 (2020)

\bibitem{kosaraju2019social}
Kosaraju, V., Sadeghian, A., Mart{\'\i}n-Mart{\'\i}n, R., Reid, I., Rezatofighi, H., Savarese, S.: Social-bigat: Multimodal trajectory forecasting using bicycle-gan and graph attention networks. Advances in Neural Information Processing Systems  \textbf{32} (2019)

\bibitem{lee2018deterministic}
Lee, J., Mansimov, E., Cho, K.: Deterministic non-autoregressive neural sequence modeling by iterative refinement. arXiv preprint arXiv:1802.06901  (2018)

\bibitem{lee2019set}
Lee, J., Lee, Y., Kim, J., Kosiorek, A., Choi, S., Teh, Y.W.: Set transformer: A framework for attention-based permutation-invariant neural networks. In: International conference on machine learning. pp. 3744--3753. PMLR (2019)

\bibitem{lee2017desire}
Lee, N., Choi, W., Vernaza, P., Choy, C.B., Torr, P.H., Chandraker, M.: Desire: Distant future prediction in dynamic scenes with interacting agents. In: Proceedings of the IEEE conference on computer vision and pattern recognition. pp. 336--345 (2017)

\bibitem{lerner2007crowds}
Lerner, A., Chrysanthou, Y., Lischinski, D.: Crowds by example. Computer graphics forum  \textbf{26}(3),  655--664 (2007)

\bibitem{liu2019naomi}
Liu, Y., Yu, R., Zheng, S., Zhan, E., Yue, Y.: Naomi: Non-autoregressive multiresolution sequence imputation. Advances in neural information processing systems  \textbf{32} (2019)

\bibitem{lucey2013representing}
Lucey, P., Bialkowski, A., Carr, P., Morgan, S., Matthews, I., Sheikh, Y.: Representing and discovering adversarial team behaviors using player roles. In: Proceedings of the IEEE Conference on Computer Vision and Pattern Recognition. pp. 2706--2713 (2013)

\bibitem{mao2020history}
Mao, W., Liu, M., Salzmann, M.: History repeats itself: Human motion prediction via motion attention. In: Computer Vision--ECCV 2020: 16th European Conference, Glasgow, UK, August 23--28, 2020, Proceedings, Part XIV 16. pp. 474--489. Springer (2020)

\bibitem{mao2019learning}
Mao, W., Liu, M., Salzmann, M., Li, H.: Learning trajectory dependencies for human motion prediction. In: Proceedings of the IEEE/CVF international conference on computer vision. pp. 9489--9497 (2019)

\bibitem{mao2023leapfrog}
Mao, W., Xu, C., Zhu, Q., Chen, S., Wang, Y.: Leapfrog diffusion model for stochastic trajectory prediction. In: Proceedings of the IEEE/CVF conference on computer vision and pattern recognition. pp. 5517--5526 (2023)

\bibitem{martinez2017human}
Martinez, J., Black, M.J., Romero, J.: On human motion prediction using recurrent neural networks. In: Proceedings of the IEEE conference on computer vision and pattern recognition. pp. 2891--2900 (2017)

\bibitem{monti2021dag}
Monti, A., Bertugli, A., Calderara, S., Cucchiara, R.: Dag-net: Double attentive graph neural network for trajectory forecasting. In: 2020 25th International Conference on Pattern Recognition (ICPR). pp. 2551--2558. IEEE (2021)

\bibitem{navarro2022social}
Navarro, I., Oh, J.: Social-patternn: Socially-aware trajectory prediction guided by motion patterns. In: 2022 IEEE/RSJ International Conference on Intelligent Robots and Systems (IROS). pp. 9859--9864. IEEE (2022)

\bibitem{ngiam2021scene}
Ngiam, J., Caine, B., Vasudevan, V., Zhang, Z., Chiang, H.T.L., Ling, J., Roelofs, R., Bewley, A., Liu, C., Venugopal, A., et~al.: Scene transformer: A unified architecture for predicting multiple agent trajectories. arXiv preprint arXiv:2106.08417  (2021)

\bibitem{omidshafiei2022multiagent}
Omidshafiei, S., Hennes, D., Garnelo, M., Wang, Z., Recasens, A., Tarassov, E., Yang, Y., Elie, R., Connor, J.T., Muller, P., et~al.: Multiagent off-screen behavior prediction in football. Scientific reports  \textbf{12}(1), ~8638 (2022)

\bibitem{pappalardo2019playerank}
Pappalardo, L., Cintia, P., Ferragina, P., Massucco, E., Pedreschi, D., Giannotti, F.: Playerank: data-driven performance evaluation and player ranking in soccer via a machine learning approach. ACM Transactions on Intelligent Systems and Technology (TIST)  \textbf{10}(5),  1--27 (2019)

\bibitem{pellegrini2009you}
Pellegrini, S., Ess, A., Schindler, K., Van~Gool, L.: You'll never walk alone: Modeling social behavior for multi-target tracking. In: 2009 IEEE 12th international conference on computer vision. pp. 261--268. IEEE (2009)

\bibitem{qi2020imitative}
Qi, M., Qin, J., Wu, Y., Yang, Y.: Imitative non-autoregressive modeling for trajectory forecasting and imputation. In: Proceedings of the IEEE/CVF Conference on Computer Vision and Pattern Recognition. pp. 12736--12745 (2020)

\bibitem{rempe2023trace}
Rempe, D., Luo, Z., Bin~Peng, X., Yuan, Y., Kitani, K., Kreis, K., Fidler, S., Litany, O.: Trace and pace: Controllable pedestrian animation via guided trajectory diffusion. In: Proceedings of the IEEE/CVF Conference on Computer Vision and Pattern Recognition. pp. 13756--13766 (2023)

\bibitem{saadatnejad2023social}
Saadatnejad, S., Gao, Y., Messaoud, K., Alahi, A.: Social-transmotion: Promptable human trajectory prediction. arXiv preprint arXiv:2312.16168  (2023)

\bibitem{sadeghian2018trajnet}
Sadeghian, A., Kosaraju, V., Gupta, A., Savarese, S., Alahi, A.: Trajnet: Towards a benchmark for human trajectory prediction. arXiv preprint  (2018)

\bibitem{salzmann2020trajectron++}
Salzmann, T., Ivanovic, B., Chakravarty, P., Pavone, M.: Trajectron++: Multi-agent generative trajectory forecasting with heterogeneous data for control. arXiv preprint arXiv:2001.03093  \textbf{2} (2020)

\bibitem{sha2017fine}
Sha, L., Lucey, P., Zheng, S., Kim, T., Yue, Y., Sridharan, S.: Fine-grained retrieval of sports plays using tree-based alignment of trajectories. arXiv preprint arXiv:1710.02255  (2017)

\bibitem{sun2019stochastic}
Sun, C., Karlsson, P., Wu, J., Tenenbaum, J.B., Murphy, K.: Stochastic prediction of multi-agent interactions from partial observations. arXiv preprint arXiv:1902.09641  (2019)

\bibitem{teranishi2022evaluation}
Teranishi, M., Tsutsui, K., Takeda, K., Fujii, K.: Evaluation of creating scoring opportunities for teammates in soccer via trajectory prediction. In: International Workshop on Machine Learning and Data Mining for Sports Analytics. pp. 53--73. Springer (2022)

\bibitem{vaswani2017attention}
Vaswani, A., Shazeer, N., Parmar, N., Uszkoreit, J., Jones, L., Gomez, A.N., Kaiser, {\L}., Polosukhin, I.: Attention is all you need. Advances in neural information processing systems  \textbf{30} (2017)

\bibitem{velivckovic2017graph}
Veli{\v{c}}kovi{\'c}, P., Cucurull, G., Casanova, A., Romero, A., Lio, P., Bengio, Y.: Graph attention networks. arXiv preprint arXiv:1710.10903  (2017)

\bibitem{vidal2022automatic}
Vidal-Codina, F., Evans, N., El~Fakir, B., Billingham, J.: Automatic event detection in football using tracking data. Sports Engineering  \textbf{25}(1), ~18 (2022)

\bibitem{wang2023tacticai}
Wang, Z., Veli{\v{c}}kovi{\'c}, P., Hennes, D., Toma{\v{s}}ev, N., Prince, L., Kaisers, M., Bachrach, Y., Elie, R., Wenliang, L.K., Piccinini, F., et~al.: Tacticai: an ai assistant for football tactics. arXiv preprint arXiv:2310.10553  (2023)

\bibitem{xie2023omnicontrol}
Xie, Y., Jampani, V., Zhong, L., Sun, D., Jiang, H.: Omnicontrol: Control any joint at any time for human motion generation. arXiv preprint arXiv:2310.08580  (2023)

\bibitem{xu2023eqmotion}
Xu, C., Tan, R.T., Tan, Y., Chen, S., Wang, Y.G., Wang, X., Wang, Y.: Eqmotion: Equivariant multi-agent motion prediction with invariant interaction reasoning. In: Proceedings of the IEEE/CVF Conference on Computer Vision and Pattern Recognition. pp. 1410--1420 (2023)

\bibitem{xu2023uncovering}
Xu, Y., Bazarjani, A., Chi, H.g., Choi, C., Fu, Y.: Uncovering the missing pattern: Unified framework towards trajectory imputation and prediction. In: Proceedings of the IEEE/CVF Conference on Computer Vision and Pattern Recognition. pp. 9632--9643 (2023)

\bibitem{yeh2019diverse}
Yeh, R.A., Schwing, A.G., Huang, J., Murphy, K.: Diverse generation for multi-agent sports games. In: Proceedings of the IEEE/CVF Conference on Computer Vision and Pattern Recognition. pp. 4610--4619 (2019)

\bibitem{zhan2018generating}
Zhan, E., Zheng, S., Yue, Y., Sha, L., Lucey, P.: Generating multi-agent trajectories using programmatic weak supervision. arXiv preprint arXiv:1803.07612  (2018)

\bibitem{zheng2016generating}
Zheng, S., Yue, Y., Hobbs, J.: Generating long-term trajectories using deep hierarchical networks. Advances in Neural Information Processing Systems  \textbf{29} (2016)

\end{thebibliography}


\begin{thebibliography}{10}
\providecommand{\url}[1]{\texttt{#1}}
\providecommand{\urlprefix}{URL }
\providecommand{\doi}[1]{https://doi.org/#1}

\bibitem{alahi2016social}
Alahi, A., Goel, K., Ramanathan, V., Robicquet, A., Fei-Fei, L., Savarese, S.: Social lstm: Human trajectory prediction in crowded spaces. In: Proceedings of the IEEE conference on computer vision and pattern recognition. pp. 961--971 (2016)

\bibitem{alcorn2021baller2vec++}
Alcorn, M.A., Nguyen, A.: baller2vec++: A look-ahead multi-entity transformer for modeling coordinated agents. arXiv preprint arXiv:2104.11980  (2021)

\bibitem{alcorn2021baller2vec}
Alcorn, M.A., Nguyen, A.: baller2vec: A multi-entity transformer for multi-agent spatiotemporal modeling. arXiv preprint arXiv:2102.03291  (2021)

\bibitem{becker2018red}
Becker, S., Hug, R., Hubner, W., Arens, M.: Red: A simple but effective baseline predictor for the trajnet benchmark. In: Proceedings of the European Conference on Computer Vision (ECCV) Workshops. pp.~0--0 (2018)

\bibitem{ding2020graph}
Ding, D., Huang, H.H.: A graph attention based approach for trajectory prediction in multi-agent sports games. arXiv preprint arXiv:2012.10531  (2020)

\bibitem{girgis2021latent}
Girgis, R., Golemo, F., Codevilla, F., Weiss, M., D'Souza, J.A., Kahou, S.E., Heide, F., Pal, C.: Latent variable sequential set transformers for joint multi-agent motion prediction. arXiv preprint arXiv:2104.00563  (2021)

\bibitem{giuliari2021transformer}
Giuliari, F., Hasan, I., Cristani, M., Galasso, F.: Transformer networks for trajectory forecasting. In: 2020 25th international conference on pattern recognition (ICPR). pp. 10335--10342. IEEE (2021)

\bibitem{glorot2010understanding}
Glorot, X., Bengio, Y.: Understanding the difficulty of training deep feedforward neural networks. In: Proceedings of the thirteenth international conference on artificial intelligence and statistics. pp. 249--256. JMLR Workshop and Conference Proceedings (2010)

\bibitem{gupta2018social}
Gupta, A., Johnson, J., Fei-Fei, L., Savarese, S., Alahi, A.: Social gan: Socially acceptable trajectories with generative adversarial networks. In: Proceedings of the IEEE conference on computer vision and pattern recognition. pp. 2255--2264 (2018)

\bibitem{huang2019stgat}
Huang, Y., Bi, H., Li, Z., Mao, T., Wang, Z.: Stgat: Modeling spatial-temporal interactions for human trajectory prediction. In: Proceedings of the IEEE/CVF international conference on computer vision. pp. 6272--6281 (2019)

\bibitem{kim2023ball}
Kim, H., Choi, H.J., Kim, C.J., Yoon, J., Ko, S.K.: Ball trajectory inference from multi-agent sports contexts using set transformer and hierarchical bi-lstm. arXiv preprint arXiv:2306.08206  (2023)

\bibitem{kingma2014adam}
Kingma, D.P., Ba, J.: Adam: A method for stochastic optimization. arXiv preprint arXiv:1412.6980  (2014)

\bibitem{lerner2007crowds}
Lerner, A., Chrysanthou, Y., Lischinski, D.: Crowds by example. Computer graphics forum  \textbf{26}(3),  655--664 (2007)

\bibitem{loshchilov2017decoupled}
Loshchilov, I., Hutter, F.: Decoupled weight decay regularization. arXiv preprint arXiv:1711.05101  (2017)

\bibitem{pellegrini2009you}
Pellegrini, S., Ess, A., Schindler, K., Van~Gool, L.: You'll never walk alone: Modeling social behavior for multi-target tracking. In: 2009 IEEE 12th international conference on computer vision. pp. 261--268. IEEE (2009)

\bibitem{saadatnejad2023social}
Saadatnejad, S., Gao, Y., Messaoud, K., Alahi, A.: Social-transmotion: Promptable human trajectory prediction. arXiv preprint arXiv:2312.16168  (2023)

\bibitem{salzmann2020trajectron++}
Salzmann, T., Ivanovic, B., Chakravarty, P., Pavone, M.: Trajectron++: Multi-agent generative trajectory forecasting with heterogeneous data for control. arXiv preprint arXiv:2001.03093  \textbf{2} (2020)

\bibitem{sun2019stochastic}
Sun, C., Karlsson, P., Wu, J., Tenenbaum, J.B., Murphy, K.: Stochastic prediction of multi-agent interactions from partial observations. arXiv preprint arXiv:1902.09641  (2019)

\bibitem{teranishi2022evaluation}
Teranishi, M., Tsutsui, K., Takeda, K., Fujii, K.: Evaluation of creating scoring opportunities for teammates in soccer via trajectory prediction. In: International Workshop on Machine Learning and Data Mining for Sports Analytics. pp. 53--73. Springer (2022)

\bibitem{wang2019translating}
Wang, Z., Liu, J.C.: Translating math formula images to latex sequences using deep neural networks with sequence-level training (2019)

\bibitem{xu2022remember}
Xu, C., Mao, W., Zhang, W., Chen, S.: Remember intentions: Retrospective-memory-based trajectory prediction. In: Proceedings of the IEEE/CVF Conference on Computer Vision and Pattern Recognition. pp. 6488--6497 (2022)

\bibitem{xu2023eqmotion}
Xu, C., Tan, R.T., Tan, Y., Chen, S., Wang, Y.G., Wang, X., Wang, Y.: Eqmotion: Equivariant multi-agent motion prediction with invariant interaction reasoning. In: Proceedings of the IEEE/CVF Conference on Computer Vision and Pattern Recognition. pp. 1410--1420 (2023)

\bibitem{xu2023uncovering}
Xu, Y., Bazarjani, A., Chi, H.g., Choi, C., Fu, Y.: Uncovering the missing pattern: Unified framework towards trajectory imputation and prediction. In: Proceedings of the IEEE/CVF Conference on Computer Vision and Pattern Recognition. pp. 9632--9643 (2023)

\bibitem{yeh2019diverse}
Yeh, R.A., Schwing, A.G., Huang, J., Murphy, K.: Diverse generation for multi-agent sports games. In: Proceedings of the IEEE/CVF Conference on Computer Vision and Pattern Recognition. pp. 4610--4619 (2019)

\end{thebibliography}
\end{document}


\title{(Supplementary) \\ TranSPORTmer: A Holistic Approach to Trajectory Understanding in Multi-Agent Sports} 

\titlerunning{TranSPORTmer}

\author{Guillem Capellera\inst{1,2}\orcidlink{0009-0006-7266-078X} \and
Luis Ferraz\inst{1}\orcidlink{0000-0001-7851-9193} \and Antonio Rubio\inst{1}\orcidlink{0000-0002-6771-8645} \and Antonio Agudo\inst{2}\orcidlink{0000-0001-6845-4998} \and Francesc Moreno-Noguer\inst{2}\orcidlink{0000-0002-8640-684X}}

\authorrunning{G.~Capellera {\em et al.}}

\institute{Kognia Sports Intelligence, Barcelona, Spain \\ \email{\{guillem.capellera,luis.ferraz,antonio.rubio\}@kogniasports.com} \and 
Institut de Robòtica i Informàtica Industrial CSIC-UPC, Barcelona, Spain
\email{\{gcapellera,aagudo,fmoreno\}@iri.upc.edu} }

\maketitle


\section{Soccer baselines}
The task of forecasting (or imputing) player trajectories based on the future movement of the ball or the opposing team has been seldom explored by state-of-the-art methods. To our knowledge, there are few baseline methods that address this aspect in forecasting tasks~\cite{yeh2019diverse,alcorn2021baller2vec++}. Therefore, we implemented these pipelines to showcase the results for the soccer dataset presented in Table 1 of the main paper. The baselines used for comparison are described as follows:

\noindent\textbf{Velocity:} As a sanity check, we adopted this baseline, projecting agent predictions linearly based on observed velocity.

\noindent\textbf{RNN:} This baseline utilizes an encoder with LSTM, employing shared weights to capture input representations of each agent, along with an MLP decoder for prediction \cite{becker2018red}.

\noindent\textbf{GRNN}: This is a non-variational version of GVRNN \cite{yeh2019diverse, sun2019stochastic}, generating trajectories without sampling. The training process involves using the ADE loss ($\mathcal{L}_\text{ADE}$). Yeh et al. \cite{yeh2019diverse} demonstrated superior results of GRNN over GVRNN in a soccer context. We use the implementation from Teranishi et al.~\cite{teranishi2022evaluation}.

\noindent\textbf{GRNN + Att}: Similar to the previous baseline but using a Graph Attention Network (GAT) instead of GNNs, inspired by \cite{ding2020graph, huang2019stgat}.

\noindent\textbf{Transformer}: Inspired by \cite{alcorn2021baller2vec, alcorn2021baller2vec++}, this baseline uses the same pipeline as ours but incorporates attention through the flattened temporal and social dimensions. It employs a 2D positional encoder~\cite{wang2019translating}, making it a non-equivariant baseline.



\section{Training procedure}
The models were trained on an NVIDIA RTX A6000 GPU for 100 epochs, using a batch size of 64 samples. We employed the AdamW optimizer \cite{kingma2014adam, loshchilov2017decoupled} with a learning rate of 0.001 and an epsilon value of $1 \times 10^{-4}$. The learning rate was reduced by a factor of 0.5 every 20 epochs. To prevent gradient explosion, we applied gradient clipping with a threshold of 5, ensuring stable optimization. Model weights were initialized using the Xavier normal distribution \cite{glorot2010understanding}. All experiments have been trained separately.

The hyperparameters for our method and were set as follows: 128-dimensional embeddings ($d$), 16 heads in each Multi-Head-Attention (MHA) module, a hidden dimension of 512 for each Set Attention Block (SAB). In \methodname~with state classification, the weight for the Cross-Entropy (CE) loss was set to $\lambda = 4$ to balance the magnitudes of the two losses. It is worth noting that without state classification, $\lambda = 0$.

\section{Basketball experiments}

\subsection{Forecasting with conditioning}
In Table 2 of the main paper, we conduct a comparative analysis between our method and forecasting-based models using the Basketball-VU dataset. The majority of these models are generative stochastic approaches primarily focused on imitative tasks. However, they are generally sub-optimal at handling a significant number of agent interactions. As a result, their ability to accurately forecast trajectories for offensive and defensive players jointly is hindered, especially when trying to model trajectories for both teams simultaneously. Consequently, these models often resort to separately modeling these trajectories. In contrast, our method can model offensive and defensive players' trajectories simultaneously using a single model.

Furthermore, our approach can generate trajectories based on the movement of the opponent players and/or the ball. This capability allows us to predict their future movements much more accurately. To illustrate this point, we would like to present an additional experiment similar to Table 1 in our main paper but focused on the basketball context. This experiment showcases the results of forecasting basketball trajectories for players or the offensive/defensive team when conditioned on the other team and/or the ball. The results of this experiment are presented in Table~\ref{tab:basketball_forec2}, demonstrating the efficacy of our model in encoding these interactions and providing more accurate results as the conditioning agents increase. It's important to note that defense predictions show significant improvement when conditioned on the offensive team alone, more so than when conditioned on the ball. This is due to the intrinsic nature of basketball, particularly in one-on-one defense situations.

\begin{table}[H]
  \centering
  \resizebox{11 cm}{!} {
  \begin{tabular}{l<{\hspace{20pt}}c <{\hspace{10pt}}c <{\hspace{10pt}}c <{\hspace{10pt}}c <{\hspace{10pt}}c <{\hspace{10pt}}c <{\hspace{10pt}}c <{\hspace{10pt}}c <{\hspace{10pt}}c <{\hspace{10pt}}c}
    \toprule
    Predict $P$ & \multicolumn{2}{c}{\textbf{Players}} & \multicolumn{4}{c}{\textbf{Offense}} & \multicolumn{4}{c}{\textbf{Defense}} \\
    \cmidrule(rr){2-3} \cmidrule(rr){4-7} \cmidrule(rr){8-11}
    Condition & None & Ball & None & Ball & Defense & Ball+Defense & None & Ball & Offense & Ball+Offense \\
    \midrule
     ADE$_P$ $\downarrow$  & 7.75 & 7.05 & 9.19 & 8.47 & 4.29 & 3.96 & 6.31 & 5.64 & 3.67 & 3.14\\
     FDE$_P$ $\downarrow$ & 11.65 & 11.13 & 14.24 & 13.78 & 7.44 & 7.05 & 9.04  & 8.48 & 5.78 & 4.76 \\
    \bottomrule
  \end{tabular}}
  \vspace{0.2cm}
  \caption{\textbf{Evaluation in Basketball-VU dataset in player trajectory forecasting with \methodname~w/o CLS.} Predictions are generated with a time horizon of 8s using a prior of 2s. All metrics are in feet.}
  \label{tab:basketball_forec2}
\end{table} 

\subsection{Unified imputation and forecasting}

\begin{figure}[H]
  \centering
  \includegraphics[trim={0 0 0 7cm}, width=0.95\columnwidth]{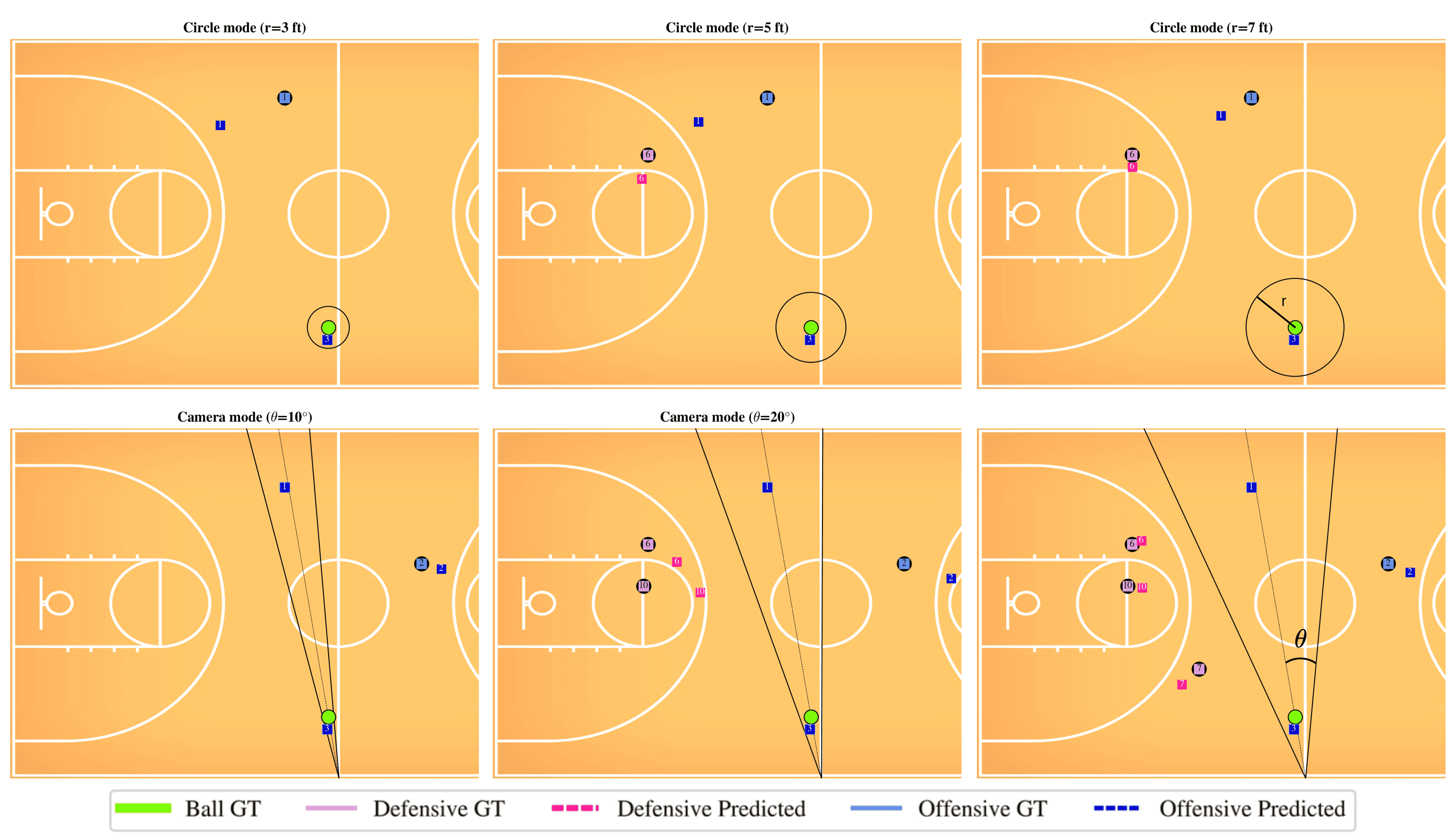}
\caption{\textbf{Initial timestep example of Basketball-TIP dataset for ``circle mode'' and ``camera mode''.} The depicted (predicted) players are those with at least one observation inside the circle/camera view during the imputation task.} 
  \label{fig:basket_ini}
\end{figure}

\begin{figure}[H]
  \centering
  \includegraphics[trim={0 0 0 7cm}, width=0.95\columnwidth]{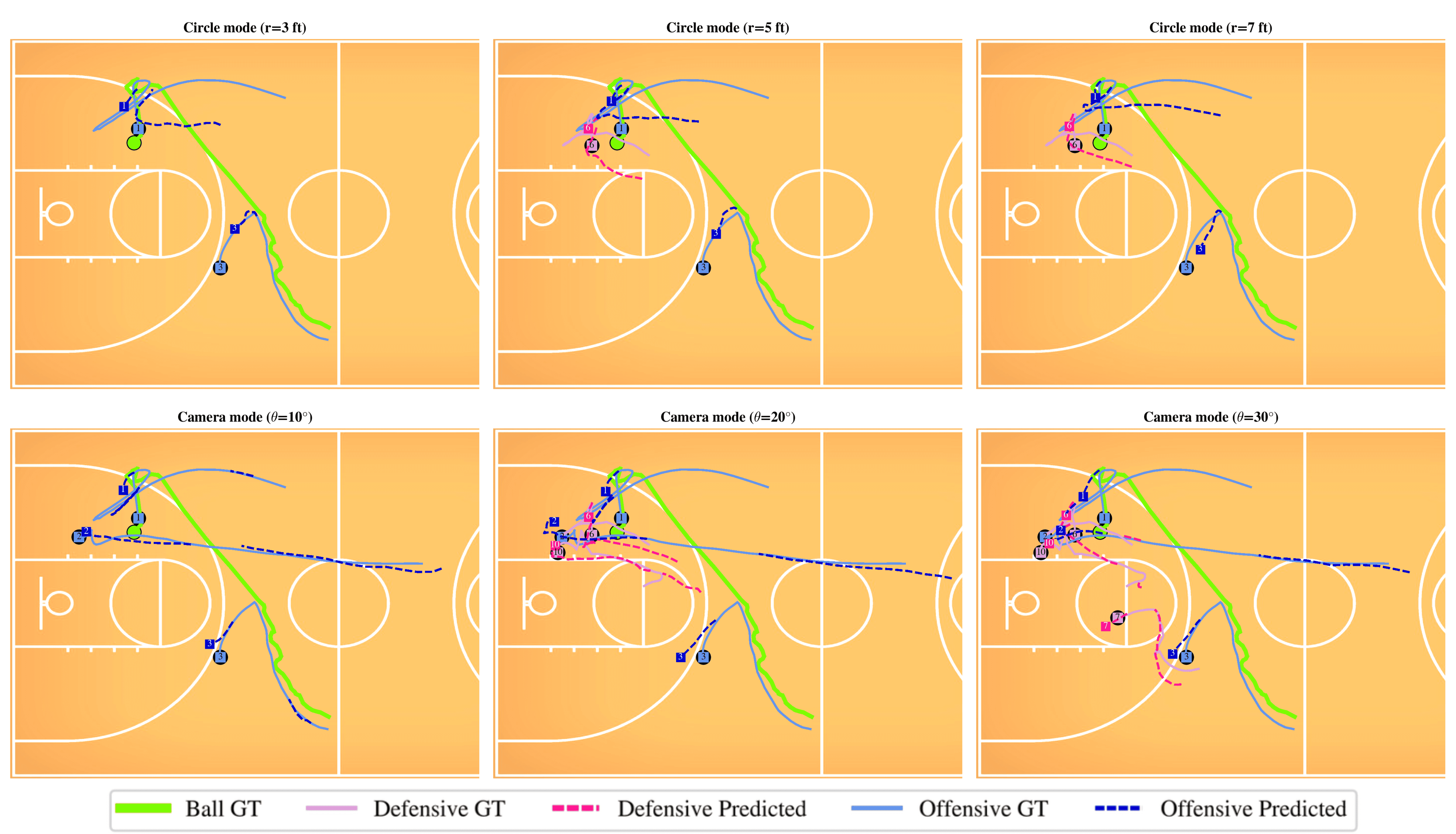}
\caption{\textbf{Qualitative evaluation in Basketball-TIP dataset.} Imputation during first 6.4s and forecasting during the subsequent 1.6s. Refer to the supplemental video to view the animated version.} 
  \label{fig:basket_fin}
\end{figure}

In Table 3 of the main paper, we present a comparison against the state-of-the-art in unified imputation and forecasting tasks using the Basketball-TIP dataset. Here, we include two figures to provide visual support for the concepts of ``circle mode'' and ``camera mode'', showing one test sample solved for each strategy.  Refer to Fig.~\ref{fig:basket_ini} to view the initial time-step of the task, where the simulated view is depicted. Refer to Fig.~\ref{fig:basket_fin} to view the same sequence solved until the final step with the predicted trajectories.

\section{Training with missing data}
In this section, we present two experiments conducted using the soccer dataset, focusing on handling missing data for goalkeepers. The first experiment involves generating the adapted dataset, enabling us to compare our approach against \textit{ballradar}~\cite{xu2023uncovering}. The second experiment assesses our model's capability to predict ball movements using the original dataset.
\subsection{Adapted dataset: goalkeepers imputation and inference}
We outline the methodology for goalkeeper trajectory imputation and inference, specifically tailored for creating the adapted dataset to facilitate a comparison between \methodname~and \textit{ballradar}~\cite{kim2023ball} in ball trajectory tasks. The \textit{ballradar} baseline relies on the positions of both goalkeepers for optimal performance in ball prediction tasks.

We initiate the process with the original soccer dataset discussed in Section 5.1 of the main paper. Since the observations are derived from optical tracking, several goalkeeper data points are missing. To address this, we initially filter and retain sequences containing at least one observation of a goalkeeper. Subsequently, we train \methodname~in goalkeeper trajectory prediction using these adapted sequences. A random mask is applied, obscuring 97\% to 100\% of the goalkeepers' observations. The goalkeepers' unavailable observations are ignored by our model using the NaN-mask during training.

\begin{figure}[H]
  \centering
  \includegraphics[width=0.8\columnwidth]{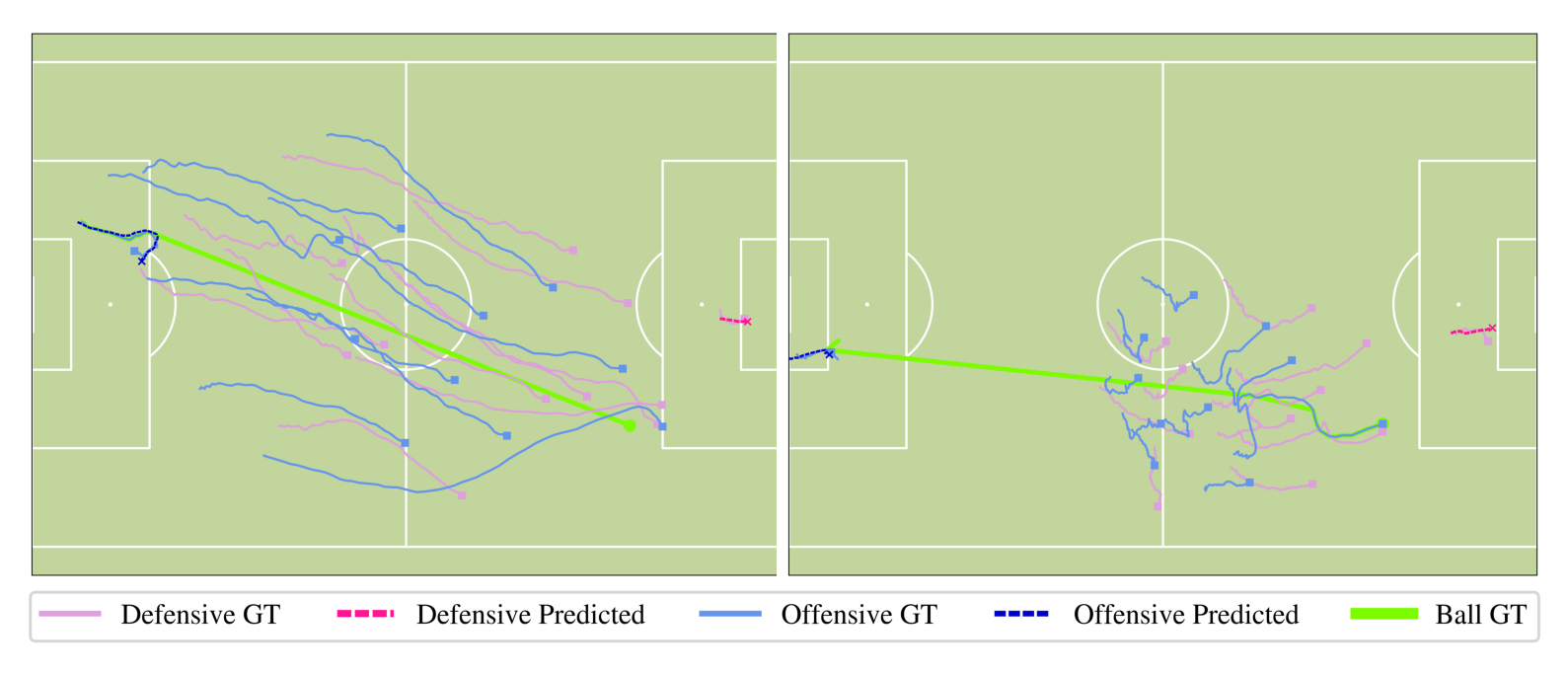}
\caption{\textbf{Goalkeepers inference through the full 9.6s sequence.} All available goalkeepers' observations are hidden.} 
  \label{fig:gkinference}
\end{figure}

The evaluation metrics for goalkeepers' trajectory inference and imputation in the test samples are presented in~\Cref{tab:gkInf1}. In the inference task, all available observations are concealed (100\% mask), and trajectory imputation is performed with 97\% of available observations hidden (97\% mask). Figure~\ref{fig:gkinference} showcases two test samples solved with all ground truth observations hidden (100\% mask). Using this trained model, we construct the adapted dataset by imputing the positions of missing observations for goalkeepers with at least one observation. In cases where one goalkeeper has no observations throughout the entire sequence, we manually set its position to a standard field position. Consequently, the new adapted dataset comprises 73,595 sequences for training, 6,628 for validation, and 5,725 for testing.

\vspace{-10mm}
\begin{table}[H]
  \centering
  \resizebox{5 cm}{!} {
  \begin{tabular}{l<{\hspace{8pt}}c<{\hspace{8pt}}c<{\hspace{8pt}}c}
    \toprule
    Task
    & Mask 
    & ADE $\downarrow$
    & MaxErr $\downarrow$ \\
    \midrule
    Inference & 100\% & 1.97 & 3.32 \\
    Imputation & 97\% & 0.82 & 1.96 \\
    \bottomrule
  \end{tabular}}
  \vspace{0.2cm}
  \caption{\textbf{Evaluation of goalkeepers' imputation and inference on 9.6s sequences.} The model utilized is \methodname~w/o CLS. All metrics are in meters.}
  \label{tab:gkInf1}
\end{table}

\subsection{Ball inference}
To ensure a fair comparison, in the main paper, we present the results of our method trained and evaluated using the adapted soccer dataset, which contains fewer sequences but includes inferred goalkeeper positions. Here, we aim to demonstrate the effectiveness of our model in predicting ball location using the original dataset, which lacks some goalkeeper observations but has approximately 10,000 (12.71\%) more sequences for training. We present the results for ball inference in Table~\ref{tab:ballinf2}, using both the adapted dataset without missing data (adapted dataset w/o missing data) and the original dataset with missing data (original dataset w missing data). It's important to note that our model can be trained with incomplete data, allowing us to train on more sequences and leading to improved results.

\begin{table}[H]
  \centering
  \resizebox{11 cm}{!} {
  \begin{tabular}{l*{5}{<{\hspace{20pt}}c}}
    \toprule
    & \multicolumn{3}{c}{\textbf{adapted dataset w/o missing data}} & \multicolumn{2}{c}{\textbf{original dataset w missing data}}  \\
    \cmidrule(rr){2-4} \cmidrule(rr){5-6}
    & ballradar (KDD'23)
    & Ours w/o CLS
    & Ours
    & Ours w/o CLS
    & Ours \\
    \midrule
     ADE $\downarrow$  & 3.89 & 2.89 & 2.71 & 2.73 & 2.57 \\
     MaxErr $\downarrow$ & 8.79 & 7.78 & 7.39 & 7.58 & 7.22 \\
     Acc (\%) $\uparrow$ & - & - & 80.84  & - & 81.64 \\
    \bottomrule
  \end{tabular}}
  \vspace{0.2cm}
  \caption{\textbf{Evaluation in ball inference in soccer.} Predictions are generated through the full 9.6s sequence. All metrics, except Acc, are in meters.}
  \label{tab:ballinf2}
\end{table} 

\section{Coarse-to-fine ablation}
\begin{figure}[H]
  \centering
  \includegraphics[width=0.7\linewidth]{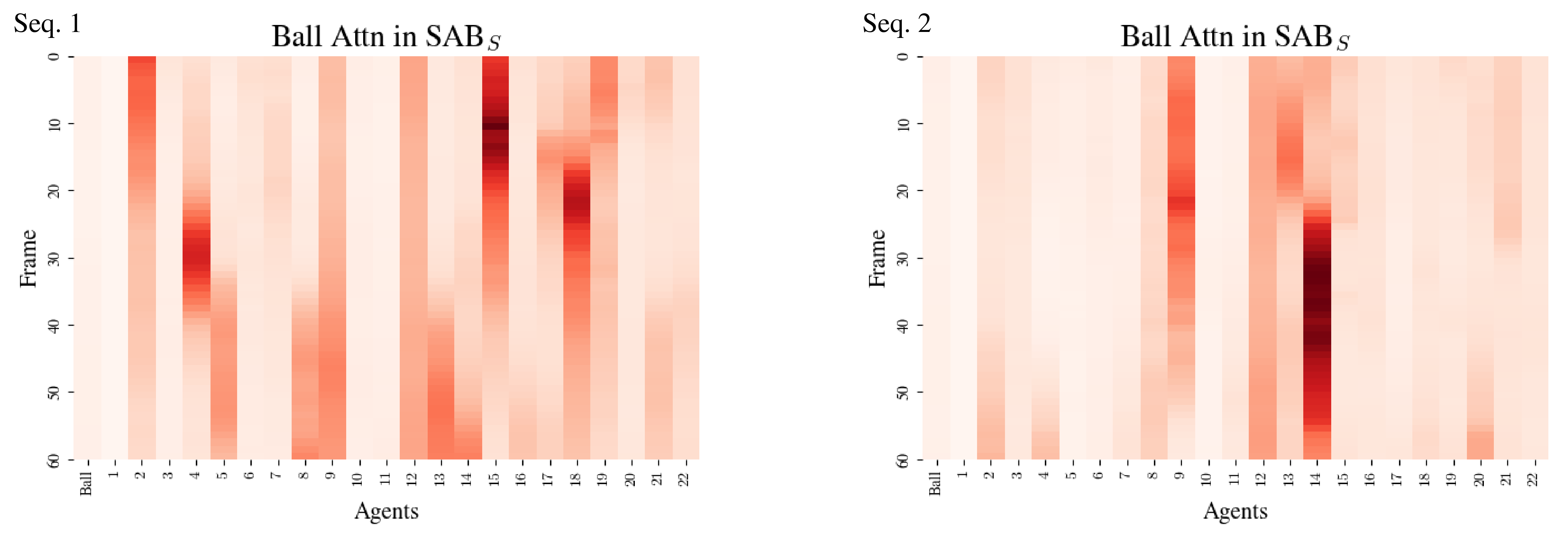}
    \vspace{-4mm}
  \caption{Results compared to Fig.4-right of the main paper using only one encoder.}
  \label{fig:attention-one-encoder}
  \vspace{-2mm}
\end{figure}

\vspace{-2mm}
In this section, we perform an ablation regarding the number of the encoders of the proposed architecture. The considered task is the ball inference and we compare \textit{Our w/o CLS}, which utilize two encoders that act as a coarse-to-fine manner, against utilizing one-single encoder and three encoders. A single encoder yelds ADE and MaxErr metrics of 4.40m and 9.83m, respectively, compared to our results of 2.71m and 7.39m (see Table 4 main paper). Figure \ref{fig:attention-one-encoder} here shows attention maps for Seq1 and Seq2 using a single encoder, exhibiting noisier focus compared to our fine encoder (see ``Ball Attn in second SAB$_S$'' in Fig.4-right main paper), leading to suboptimal results. Using three encoders fails to converge.
\vspace{-5mm}

\section{Pedestrian forecasting}
\vspace{-2mm}
Although analyzing urban pedestrian scenes is outside the main scope of this paper, for completeness we have included an experiment with the benchmark dataset ETH-UCY~\cite{pellegrini2009you, lerner2007crowds}. This dataset comprises five different subsets: ETH, Hotel, Univ, Zara1, and Zara2. We follow the established convention of leave-one-out training~\cite{gupta2018social} and employ the task of forecasting 12 future time-steps based on 8 preceding time-steps, with a frame rate of 2.5Hz. The results of our experiment against deterministic state-of-the-art models are presented in Table~\ref{tab:pedestrian}. It is worth pointing out that our approach is on a par with the most recent architectures, many of which are specifically tailored for pedestrian contexts, and achieves strong results in three subsets (ETH, Zara1, Zara2) and on Average. We achieve a 4.3\% improvement in ADE on the ETH subset.

\begin{table}[H]
\vspace{-5mm}
  \centering
  \scriptsize	
  \resizebox{12 cm}{!}{
  \begin{tabular}{@{} l @{\hspace{12pt}} c @{\hspace{12pt}} c @{\hspace{12pt}} c @{\hspace{12pt}} c @{\hspace{12pt}} c @{\hspace{12pt}} c @{\hspace{12pt}} c}
    \toprule
    Model
    &  & ETH & Hotel & Univ & Zara1 & Zara2 & Average \\
    \midrule
    S-LSTM\cite{alahi2016social} & CVPR'16 &  1.09/2.35 & 0.79/1.76 & 0.67/1.40 & 0.47/1.00 & 0.56/1.17 & 0.72/1.54\\
    SGAN-ind\cite{gupta2018social} & CVPR'18 & 1.13/2.21 & 1.01/2.18 & 0.60/1.28 & 0.42/0.91 & 0.52/1.11 & 0.74/1.54 \\
    TransF\cite{giuliari2021transformer} & ICPR'20 & 1.03/2.10 & 0.36/0.71 & \underline{0.53}/1.32 & 0.44/1.00 & 0.34/0.76 & 0.54/1.17 \\
    Trajectron++\cite{salzmann2020trajectron++} & ECCV'20 & 1.02/2.00 & 0.33/0.62 & \underline{0.53}/1.19 & 0.44/0.99 & 0.32/0.73 & 0.53/1.11   \\
    MemoNet\cite{xu2022remember} & CVPR'22 & 1.00/2.08 & 0.35/0.67 & 0.55/1.19 & 0.46/1.00 & 0.32/0.82 & 0.55/1.15    \\
    Autobots\cite{girgis2021latent} & ICLR'22 & 1.02/1.89 & \underline{0.32}/\underline{0.60} & 0.54/1.16 & 0.41/0.89 & 0.32/0.71 & 0.52/\underline{1.05}\\
    EqMotion\cite{xu2023eqmotion} & CVPR'23 & 0.96/1.92 & \textbf{0.30}/\textbf{0.58} & \textbf{0.50}/\textbf{1.10} & \textbf{0.39}/\textbf{0.86} & \textbf{0.30}/\textbf{0.68} & \textbf{0.49}/\textbf{1.03} \\
    Social-Transmotion\cite{saadatnejad2023social} & ICLR'24 & \underline{0.93}/\textbf{1.81} & \underline{0.32}/\underline{0.60} & 0.54/\underline{1.16} & 0.42/0.90 & 0.32/0.70 & \underline{0.51}/\textbf{1.03}  \\
    \midrule
    Our w/o CLS & & \textbf{0.89}/\underline{1.87} & 0.36/0.73 & 0.57/1.22 & \underline{0.40}/\underline{0.87} & \underline{0.31}/\underline{0.69} & \underline{0.51}/1.08\\
    \bottomrule
  \end{tabular}}
  \vspace{0.2cm}
  \caption{\textbf{Evaluation on ETH-UCY dataset in pedestrian forecasting (ADE/FDE).} The observation is performed during 8 time-steps (3.2s) while the forecasting is performed during the subsequent 12 time-steps (4.8s). Results are extracted from previous works~\cite{xu2023eqmotion,saadatnejad2023social}.}
  \label{tab:pedestrian}
\end{table} 

%
%
\bibliographystyle{splncs04}
\bibliography{main}